\documentclass[11pt,letterpaper]{article}

\usepackage[margin=1in]{geometry}
\usepackage[T1]{fontenc}
\usepackage[utf8]{inputenc}
\usepackage{textcomp}
\usepackage{palatino,helvet}
\usepackage[table,xcdraw,dvipsnames]{xcolor}
\usepackage{natbib}
\setcitestyle{aysep={}}
\renewcommand{\cite}{\citep}

\usepackage{epsfig}
\usepackage{amsfonts,amssymb}
\usepackage[fleqn]{amsmath}
\usepackage{amsthm}
\usepackage{url}
\usepackage[colorlinks=true,linkcolor=blue!50!black,citecolor=blue!50!black,urlcolor=blue!50!black]{hyperref}

\newtheoremstyle{break}%
 {}{}%
 {}{}%
 {\bfseries}{}%
 {\newline}{}
\theoremstyle{break}

\newtheorem{definition}{Definition}

\usepackage{multirow}
\usepackage{fancyvrb}
\VerbatimFootnotes
\usepackage{float}
\usepackage{tikz}
\usepackage{graphicx}
\usepackage{enumitem}
\usepackage{mathtools}
\usepackage{makecell}
\usepackage{tcolorbox}

\usepackage{setspace}
\usepackage{booktabs}
\usepackage{caption}
\usepackage{subcaption}
\usepackage{lscape}

\usepackage[normalem]{ulem}

\usepackage{adjustbox}

\usepackage{tcolorbox}
\tcbuselibrary{theorems}

\begin{document}

\title{Linguistic Features for Interpretable Textual Entailment}

\author{%
David Torres-Moreno$^{1}$\thanks{The author would like to thank CONACYT MEXICO for the grant awarded to carry out this work.},
Jorge Hermosillo-Valadez$^{1}$\thanks{Corresponding author},
Asela Reig-Alamillo$^{2}$\\[1.5ex]
\small $^{1}$Centro de Investigaci\'on en Ciencias, Universidad Aut\'onoma del Estado de Morelos\\
\small \texttt{david.torres@uaem.mx}, \texttt{jhermosillo@uaem.mx}\\[0.8ex]
\small $^{2}$Centro Interdisciplinario de Investigaci\'on en Humanidades, Universidad Aut\'onoma del Estado de Morelos\\
\small \texttt{assela.reig@uaem.mx}
}
\date{}

\maketitle

\begin{abstract}
Despite the success of neural models in natural language processing, their black-box nature limits interpretability and conceals the linguistic phenomena underlying their predictions. We present SLITE, an explainable hybrid model for Recognizing Textual Entailment that integrates two complementary layers of semantic analysis: a structural-relational layer, based on semantic compatibility and incompatibility between compositional entities, and a distributional-informational layer, based on structured patterns of information change between embedding-based representations of the premise and the hypothesis. We propose 17 features that combine entity-level semantic relations, polarity-sensitive lexical matching, and alignment measures over semantic sub-representations of the similarity matrix, including measures based on entropy and transfer entropy. A logistic regression trained on these features achieves an accuracy of 83\% on three-class SICK and 96\% on SICK-CE, outperforming IsoLex by 4 percentage points and falling within 2 percentage points of RoBERTa with a fraction of its computational complexity. Ablation studies and SHAP analysis confirm that structural-relational features are the primary drivers of classification, while distributional-informational features provide essential complementary contributions, particularly for detecting neutrality and contradiction. Our results demonstrate that further exploration of hybrid approaches is a viable and scientifically productive alternative to massive neural architectures, and we hope they will strengthen the dialogue between linguistic theory and computational modeling of inference.

\end{abstract}

\section{Introduction}
\label{sec:introduccion}

In the current generation of Artificial Intelligence systems, linguistic task modeling seeks to capture the intrinsic complexity of human language, marked by lexical, structural, and semantic variability and ambiguity. Within this challenge, Recognition of Textual Entailment (RTE) or Natural Language Inference (NLI) is a fundamental pillar for evaluating language comprehension, challenging computational systems to determine whether the meaning of a hypothesis ($H$) can be derived from a premise ($P$) \citep{dagan2006}.

To make this concrete, consider the following pair of sentences:
\begin{itemize}
    \item [$1.$] A blonde woman is walking ($P$).
    \item [$2.$] A person is walking ($H$).
\end{itemize}
Humans are able to judge whether the second sentence (the Hypothesis) follows rationally from the first (the Premise) or, in other words, whether, given the truth of P, H would be rationally judged to be true. This interpretive task, routine in language processing, involves semantic comprehension and inferential processes sometimes supported by world knowledge, and poses a non-trivial challenge for Natural Language Processing (NLP).

The recent trend in the scientific community has prioritized the development of massive neural architectures, such as pre-trained transformers. Models such as BERT, RoBERTa, and their successors dominate standard benchmarks, achieving accuracies that often exceed 95\% on datasets such as SuperGLUE \citep{Wang2019SuperGLUEAS}. However, these systems operate as black boxes that, despite their high performance, lack transparency in decision-making, ignore specific linguistic phenomena, and cannot account for their predictions in terms of interpretable linguistic rules or reasoning chains \citep{souza2025hybrid, YANG2023103245}.
Adding to this concern is evidence that transformer models tend to exploit superficial artifacts in the datasets rather than acquiring robust inferential mechanisms \citep{gururangan-etal-2018-annotation}. Even large-scale benchmarks such as SNLI \citep{bowman2015large} have been questioned as adequate measures of inferential competence, suggesting that progress in NLI requires not only greater accuracy but also a clearer explanation of what models actually learn.

NLI holds a prominent place in computational linguistics. It serves both as a performance benchmark in NLP and as a window into the mechanisms and processes that underpin it. This duality makes interpretability a methodological imperative. A system capable of classifying pairs of inferences without explicitly stating the linguistic foundations of its decisions contributes little to the study of inference as a phenomenon and lacks a solid basis for diagnosing systematic errors. Beyond scientific requirements, traceability is a pragmatic necessity in critical domains such as legal reasoning, clinical medicine, or fact-checking, where the validity of a conclusion depends on its ability to be audited. In these contexts, a model’s reliability lies not only in its accuracy but also in the traceability of its inferential process.

The quest for explainability has taken two forms: \textit{post-hoc} interpretability, which involves applying explanation methods to an opaque, pre-trained model, and \textit{inherent} interpretability, in which the decision logic is transparent by design \citep{rudin2019, bukart2021}. Post-hoc attribution methods have demonstrated considerable utility for probing what information neural representations encode
\citep{belinkov2022probing, ferrando2024primer}; however, when applied to opaque architectures,
they produce approximations of model behavior rather than direct accounts of it,
and the fidelity of those approximations cannot be independently verified against
the underlying decision process \citep{NEURIPS2023_89beb2a3}. This limitation is of great relevance to the understanding and explanation of automatic inference mechanisms. When the inference chain is the primary object of study, a model whose decision logic is transparent by design offers a qualitatively different type of evidence than one whose decisions are reconstructed a posteriori.

There have been notable efforts to restore the interpretability of distributed representations through well-founded geometric and semantic constraints \citep{iwamoto-etal-2021-polar, engler-etal-2022-sensepolar}. Still, while embeddings are highly effective at capturing statistical patterns of usage in similar contexts, their interpretability is restricted to the representational level and does not extend to the inferential process. From a linguistic perspective, entailment judgments are known to depend crucially on lexical-semantic relations: in our example, the human inference rests on the relationship between \textit{woman} and \textit{person}. The link between such relations and propositional entailment is well established in formal and lexical semantics \citep{lyons1977semantics, cruse2004meaning, hurford2007semantics, jeffries1998meaning} and has been placed at the core of approaches such as Natural Logic \citep{maccartney2009}. Consequently, the nature of RTE demands a qualitative leap: implication judgments must not stem from statistical proximity, but from the structural relations between their components. Integrating distributional representations with structured symbolic knowledge—hypernymy, merony, synonymy, and antonymy—allows us to operationalize how these relationships dictate the value of the inference, thereby extending explainability from the representation to the inferential mechanism itself.

The need for structured symbolic knowledge is accentuated by the intrinsic limitations of similarity measures in geometric spaces. \citet{santus2014} demonstrate that symmetric measures, such as cosine similarity, are inadequate for capturing asymmetric semantic relations, such as hypernymy (\textit{dog} → \textit{animal}). Although distributional spaces encode semantic generality—since hypernyms typically occupy more diffuse contexts with higher entropy than their hyponyms \citep{santus2014}—their architecture is not designed to model intrinsic semantic properties, but rather patterns of statistical co-occurrence, regardless of the knowledge resources used to enrich them. Similarly, \citet{amigo2022} demonstrate, through their Information Theory-based Compositional Distributional Semantics (ICDS) framework, that models such as BERT or GPT suffer from a \textit{representation degradation problem}: contextual representations, being optimized for language modeling objectives, often sacrifice the geometric properties required for a reliable similarity measurement.

This convergence of constraints is driving the development of hybrid methods that, without compromising performance competitiveness, operate under principles of linguistic rigor and offer inherent interpretability. Recent frameworks such as XTE and IsoLex \citep{vivianXTE, souza2025hybrid} have demonstrated the potential of these approaches by integrating statistical filters, lexical resources (e.g., WordNet), and embedding similarity into serialized architectures. However, despite their transparency and achieving performance close to the state of the art, their modular design suffers from an intrinsic vulnerability to error propagation and the rigidity of their intermediate representations. This is particularly critical in the dependency on Subject-Verb-Object (SVO) triplets, which are insufficient to capture the structural complexity and diversity of syntactic configurations characteristic of natural language.

We therefore propose a framework that operationalizes
textual entailment in a non-serialized manner as an analysis of two complementary dimensions. First, we propose a \emph{Structural-Relational} dimension of analysis. We construct $P$ and $H$ entities ---head concepts with their attributes--- in order to identify semantic compatibility relationships for the RTE task. Based on this analysis, we propose relational measures and a Jaro measure modified and extended using semantic knowledge for the RTE task. This measure is based on an analysis of the structure of $P$ and $H$ in order to establish links between their lexical units, taking into account negations (polarity) that modify their meaning and the correspondence of the link.

 Second, we propose a \emph{Distributional-Informational} dimension of analysis. We analyze structured patterns of information change between the distributional representations of $P$ and $H$. Beyond cosine similarity, we construct sub-representations of the entity and lexical similarity matrices by selectively removing rows and columns corresponding to specific semantic groups. Alignment measures over the resulting submatrices quantify the residual coherence of the $P$-$H$ relationship when certain relation types are absent, operationalizing information change as a structural property of the embedding-based semantic alignment. Entropy-based measures---including transfer entropy---provide a complementary view of the same phenomenon at the distributional level, quantifying directional uncertainty reduction from $P$ to $H$.

Our work presents the following contributions:

\begin{itemize}

\item \textbf{A new methodological framework for
explainable RTE.} We propose two complementary
layers of semantic analysis, Structural-Relational
and Distributional-Informational, that jointly
operationalize textual entailment as a directed
semantic coverage and structured pattern of
information change between $P$ and $H$.
The Structural-Relational layer grounds inference
in notions of semantic compatibility and
incompatibility between compositional entities,
leveraging structured semantic knowledge from
an external knowledge resource to identify
equivalence, opposition, specificity, and
no-relation between the elements of $P$ and $H$.
The Distributional-Informational layer complements
this by capturing residual coherence and
directional information flow between their
embedding-based representations, analyzing
how information changes when specific semantic
relations are present or absent.

\item \textbf{An efficient and inherently interpretable
hybrid model.} We propose a set of 17 features
organized across the two layers described above
that, together with a logistic regression classifier,
achieve 96\% accuracy on the SICK-CE test set and
83\% on the three-class SICK dataset. This result
surpasses the hybrid-explainable IsoLex model by
4 percentage points and comes within 2 points of
RoBERTa, with a fraction of its computational
complexity, demonstrating that interpretability
and near-state-of-the-art performance need not
be in tension.

\item \textbf{A comprehensive explainability analysis using SHAP.} We apply SHAP to break down each decision made by the model, revealing that the main factors determining classification are Structural-Relational features, while Distributional-Informational features make specific but complementary contributions, particularly in the detection of contradictions.

\end{itemize}

The paper is structured as follows:
Section~\ref{sec:trabajo_relacionado} reviews
previous work and positions our contribution;
Section~\ref{sec:notacion} introduces the
representational primitives common to both layers;
Section~\ref{sec:caracteristicas} describes the
SLITE feature set and classification model;
Section~\ref{sec:experimentos} presents the
experiments, results, and analysis; and
Section~\ref{sec:conclusion} discusses implications,
limitations, and future research directions.

\section{Related work and contribution}
\label{sec:trabajo_relacionado}

The landscape of RTE research has evolved from rule-based and lexical overlap methods toward massive neural architectures, with an emerging line of hybrid approaches seeking to recover interpretability without sacrificing competitive performance. This section traces that evolution and situates our contribution within this landscape.

\subsection{Datasets and Task Evolution}
\label{subsec:datasets_evolucion}

The task of RTE arose with the PASCAL challenges, posed as a binary entailment problem \citep{dagan2006}. A qualitative leap was made with large-scale annotated datasets such as SNLI \citep{bowman2015large} and MultiNLI \citep{williams-etal-2018-broad}, which introduced the \textit{neutral} category, broadening the spectrum of reasoning required. To specifically evaluate compositional semantics, the SICK (\textit{Sentences Involving Compositional Knowledge}) corpus \citep{marelli-etal-2014-sick} was created, characterized by a controlled design that normalizes verb tenses and avoids idioms or complex named entities. Its SICK-CE subset, containing only entailment and contradiction pairs \citep{souza2025hybrid}, has become an ideal testbed for evaluating models' ability to distinguish these core relations without the added ambiguity of neutrality. Other specialized corpora, such as SciTail \citep{Khot2018SciTaiLAT}, derived from science questions, and the SuperGLUE diagnostic sets \citep{Wang2019SuperGLUEAS}, have continued to push the complexity and diversity of the task.

Pioneering methods were based on measures of lexical overlap and superficial similarity, which were quickly limited by phenomena such as negation or pragmatic implication \citep{ santus2014}.
Natural Logic \citep{maccartney-manning-2009-extended} represented a significant conceptual advance by enabling formal inference directly over
surface linguistic structures through a monotonicity calculus and lexical alignment rules. These models are intrinsically interpretable: every inference step corresponds to an explicit, human-readable operation over the logical form of the sentence.
Their coverage, however, is bounded by the expressiveness of the monotonicity calculus and the completeness of the lexical alignment rules. \citet{hu-etal-2020-monalog} report that MonaLog, a modern lightweight implementation of the Natural Logic paradigm, achieves 77.2\% accuracy on SICK---a result that is competitive within the symbolic paradigm, though one that highlights the
performance ceiling that monotonicity-based systems tend to approach on large-scale benchmarks. Critically, this gap
persists and widens on the Logic and Commonsense subsets of SICK \citep{kalouli-etal-2023-curing}, which were specifically designed to target inferential phenomena that Natural Logic should handle well (see Section \ref{subsec:resultados_cuantitativos} for a detailed comparison), suggesting that the limitation is not merely one of
scalability but of inferential coverage.

The current dominant paradigm consists of neural models, first with architectures based on LSTM and attention mechanisms \citep{rocktaschel2016reasoning,parikh-etal-2016-decomposable,zhao-etal-2016-textual}, and subsequently with the fine-tuning of pre-trained transformers such as BERT \citep{devlin2019bert} and RoBERTa \citep{liu2019roberta}. These models have established the state of the art (SOTA) in virtually all RTE benchmarks, achieving, for example, 98\% accuracy on SICK-CE in the case of RoBERTa \citep{souza2025hybrid}. However, their black-box nature and tendency to learn superficial biases from the data rather than deep reasoning have prompted growing criticism and the search for more explainable alternatives \citep{kalouli-etal-2023-curing}.

An emerging area of research seeks to combine the predictive power of statistical approaches with the transparency of symbolic methods.
One proposal is the XTE (Explainable Text Entailment) model \citep{vivianXTE}, designed to overcome the opacity and rigidity of traditional systems. Instead of operating as a black box, XTE integrates specialized components (one for syntactic analysis and another for semantic analysis), which are activated on a case-by-case basis. Its architecture allows it to generate justifications in natural language for each decision and base its reasoning on external structured knowledge, such as dictionary definitions, thus reducing dependence on biases present in the training data. This hybrid approach proved its effectiveness, achieving an F1 score of 0.59 in an adaptation of the SICK dataset, where the contradiction and neutral categories were grouped under a single non-entailment label.

A parallel line of research has explored the potential of Information Theory as a foundation for semantic representation and inference. \citet{santus2014} demonstrated that distributional entropy can capture asymmetric semantic relations such as hypernymy that symmetric measures like cosine similarity cannot, grounding semantic generality in the diversity of contextual distributions. \citet{amigo2022} extended this perspective through their ICDS framework, showing that information-theoretic composition can provide a more principled account of meaning than geometric operations over embedding spaces. More recently, \citet{staliunaite-vlachos-2025-uncertain} have pointed to the potential of semantic entropy for modeling ambiguity and label variation in RTE. Together, these works suggest that Information Theory offers tools that are both formally grounded and linguistically interpretable---properties that, to our knowledge, have not yet been exploited in the form of directional information flow features for RTE classification.

The most direct precedent for our work is IsoLex \citep{souza2025hybrid}, which serves as our primary baseline. Its serialized architecture, however, presents two limitations that motivate our design. First, the pipeline structure is prone to cascading errors, where missteps at intermediate stages propagate uncorrected to the final classifier. Second, and more fundamentally, the reliance on Subject-Verb-Object (SVO) triplets for structural representation constrains syntactic coverage in ways well-documented in the information extraction literature \citep[inter alia]{Banko2007}. Complex constructions such as relative clauses (\textit{the man who was running}) and secondary predication (\textit{she left angry}) are frequently misanalyzed or discarded by SVO extraction pipelines. While passive alternations (\textit{the instrument was put away by the woman}) pose challenges for both SVO and dependency-based representations, the latter preserves the essential argument structure required for semantic analysis rather than silently omitting it. Given that such constructions are prevalent in naturally occurring sentence pairs, a representation that fails to capture them systematically underrepresents the structural evidence necessary for accurate entailment judgments. Unlike IsoLex, which relies on WordNet for lexical relation extraction, our approach uses ConceptNet \citep{speer2018conceptnet55openmultilingual}, whose richer relational inventory and compatibility with dense vector representations make it better suited to our non-serialized architecture.

Another line of work is represented by the Semantic Knowledge Abstraction (SKA) proposal \citep{TORRESMORENO2026114825}.
The SKA framework reconfigures the lexical-semantic relationships between $P$ and $H$ into four functional categories---equivalence/generality, opposition, specificity, and no relation---that align with the logical classes of entailment, contradiction, and neutrality. This categorical representation reduces reliance on surface-level lexical patterns and provides a structured basis for semantic compatibility judgments. In the present work, we reformulate these notions within a formal dependency-based framework and extend their operationality as the symbolic backbone of our feature set, combining them for the first time with information-theoretic measures of directional information flow to produce a unified, non-serialized representation of the $P$-$H$ relationship.

These hybrid approaches collectively demonstrate
that the perceived trade-off between performance
and interpretability in RTE is far from an inherent
necessity; rather, it reflects the current dominance
of opaque neural architectures, which has obscured
the potential for high-performing, explainable models.
Yet no existing model simultaneously addresses the
structural limitations of serialized pipelines, the
geometric constraints of cosine-based similarity,
and the requirement for directional information flow.
Our proposed architecture addresses each of these
limitations directly through its two complementary
layers of analysis: the Structural-Relational layer
overcomes the rigidity of serialized pipelines and
cosine-based representations, while the
Distributional-Informational layer operationalizes
directional information flow as a structural property
of the $P$-$H$ semantic alignment.

\subsection{Contribution}
\label{subsec:nuestra_contribucion}

Our proposal differs from previous work in two key
conceptual and methodological aspects:

\begin{enumerate}
    \item \textbf{Structured Semantic Analysis.}
    Unlike IsoLex and other models that rely on
    rigid extraction of SVO triplets, vulnerable
    to syntactic variation, we perform an analysis
    of entities and their attributes based on
    syntactic dependencies, which we term the
    \emph{Structural-Relational} layer of analysis.
    We categorize the entities within $H$ into four
    disjoint groups based on their semantic
    relationship with $P$: equivalence/generality,
    opposition, specificity, and no relationship.
    Based on this analysis, we propose two types
    of measures: relational features that quantify
    the distribution and interaction of semantic
    groups, and a semantically extended Jaro measure
    that establishes links between the lexical units
    of $P$ and $H$ using ConceptNet relations,
    incorporates polarity-sensitive matching to
    account for negation, and applies directional
    penalties reflecting the type and correspondence
    of each link. Together, these measures
    operationalize directed semantic coverage at
    a level of granularity that pure distributional
    approaches cannot encode directly.

    \item \textbf{Quantification of Information
    Flow.} While most models, including IsoLex,
    rely on cosine similarity as their primary
    measure of semantic proximity, our
    \emph{Distributional-Informational} layer goes
    beyond this by analyzing structured patterns
    of information change between the distributional
    representations of $P$ and $H$. Prior work has
    established the relevance of distributional
    entropy for capturing asymmetric semantic
    relations \citep{santus2014}, and
    information-theoretic composition has been
    proposed as a principled alternative to
    geometric operations over embedding spaces
    \citep{amigo2022}; however, these approaches
    operate at the level of individual word
    distributions rather than at the level of
    structured semantic sub-representations. Our
    proposal extends this line of work by
    constructing sub-representations of the
    similarity matrices through selective removal
    of rows and columns corresponding to specific
    semantic groups, and computing alignment and
    entropy-based measures, including transfer
    entropy \citep{PhysRevLett.85.461}, over the
    resulting submatrices, operationalizing
    information change as a structural property
    of the $P$-$H$ semantic alignment.
\end{enumerate}

Both contributions share a fundamental requirement
for the underlying knowledge resource: it must
support simultaneous access at the symbolic level
and the distributional level. WordNet
\citep{Fellbaum1998}, employed by IsoLex, is
structured primarily around taxonomic hierarchies,
offering limited coverage of the non-taxonomic
relations required by our feature set. ConceptNet
\citep{speer2018conceptnet55openmultilingual},
by contrast, provides a richer relational inventory
and grounds its concepts in Numberbatch embeddings,
enabling unified symbolic and distributional
operations within a single resource. A detailed
justification of this choice is provided in
Section~\ref{sec:notacion}.

\section{Representational Framework: Entities, Matrices,
and Semantic Categories}
\label{sec:notacion}
The framework proposed here operates across two
complementary layers of semantic analysis. The
Structural-Relational layer decomposes sentences
into compositional entities and their attributes,
which are then categorized according to the
lexical-semantic relations they bear to one
another---relations whose connection to entailment
judgments is grounded in classical lexical and
formal semantics, as we discuss in
Section~\ref{subsec:categorizacion}. The
Distributional-Informational layer constructs
similarity matrices over word embeddings as the
basis for computing alignment and entropy-based
measures of information change between $P$ and
$H$ (see Figure~\ref{fig:pipeline}). The following
subsections introduce the representational
primitives common to both layers: the entity and
lexical representations, the similarity matrices,
and the semantic categorization of relations.

ConceptNet \citep{speer2018conceptnet55openmultilingual} is selected as the primary knowledge resource because it provides a broader range of semantic relations than WordNet, including meronymy, antonymy, and commonsense associations, which are essential for modeling entailment and contradiction in RTE tasks. In addition, its multilingual graph structure and integration with Numberbatch embeddings \citep{speer2018conceptnet55openmultilingual} enable both symbolic reasoning over explicit relations and distributional analysis through dense vector representations within a unified framework. This combination is particularly suitable for our non-serialized architecture, where symbolic and distributional analyses must operate over compatible representations.

\subsection{Word Processing and Representation}
\label{subsec:procesamiento}

Each pair of sentences (premise $P$ and hypothesis $H$) is processed using the \texttt{spaCy} pipeline (model \texttt{en\_core\_web\_md}) enriched with ConceptNet Numberbatch embeddings\footnote{https://github.com/commonsense/conceptnet-numberbatch}.
This process generates two structured representations:

\begin{itemize}
    \item \textbf{Lexical Representation:} All words in $P$ and $H$ (excluding punctuation marks) are lemmatized to obtain flat lists of lemmas:
    $$
    \mathcal{L}_P = [w^P_1, \dots, w^P_p]
    $$
    $$
    \mathcal{L}_H = [w^H_1, \dots, w^H_q],
    $$
    where $w^\text{x}_i$ is a lemmatized word in either $P$ or $H$.

    \item \textbf{Structure of Entities with Attributes:} Through dependency analysis, the main entities (nouns and verbs) of each sentence and their respective attributes (adjectives, adverbs, prepositions) are extracted. The result is an associative data structure, composed of tuples, where the first element of each tuple is a lemmatized entity, and the subsequent optional elements of the tuple are their corresponding lemmatized attributes. Thus, we define two sets of entities of $P$ and entities of $H$ as follows:
    \[
    \mathcal{E}_P = \{ e^P_i=(w_i^{\epsilon _P}, [;w^{a_P}_1,\dots,w^{a_P}_r]) | \; w^\epsilon_i,w^{a_P}_r  \in \mathcal{L}_P,\,1\leq i\leq m,\text{ and } r=1,2,...\}
    \]
    \[
    \mathcal{E}_H = \{ e^H_j=(w^{\epsilon_H}_j, [;w^{a_H}_1,\dots,w^{a_H}_r]) | \; w^\epsilon_j,w^{a_H}_r \in \mathcal{L}_H,\,1\leq j\leq n,r=1,2,...\}
    \]
    Example:
        $P$: ``\textit{a young player is throwing the red ball}'', we obtain
        $$\mathcal{E}_P = \{(\text{`player'},[\text{`young'}]), (\text{`throw'},[\text{`be'}]), (\text{`ball'},[\text{`red'}])\}.$$
\end{itemize}

\subsection{Construction of Similarity Matrices}
\label{subsec:matrices}

Based on the above representations, we construct two cosine similarity matrices that capture semantic alignment at different levels of granularity.

\subsubsection{Entity Similarity Matrix  ($\mathbf{M}_e$)}
\label{subsubsec:matriz_entidades}
For pure entities (not including attributes), we compute the cosine similarity between the embeddings of all entities in $P$ and $H$ respectively. We recall the definition of the cosine similarity measure:
$$
\text{Cosine Similarity }:= \sigma(a,b)= \cos(\theta)=\frac{a\cdot b}{\lVert a \rVert \lVert b \rVert}
$$
where $\theta$ is the angle subtended between vectors $a$ and $b$. Therefore, cosine similarity is defined in the range $[-1,1]$.

Let $\mathbf{v}(\cdot)$ be the embedding of some entity. We define the matrix $\mathbf{M}_e \in [0,1]^{m \times n}$, where $m = |\mathcal{E}_P|$, $n = |\mathcal{E}_H|$:
\[
(\mathbf{M}_e)_{ij} = \sigma\left(\mathbf{v}(w^\epsilon_i), \mathbf{v}(w^\epsilon_j)\right),\;0\leq i\leq m\;\text{ and }0\leq j\leq n,
\]
$w^\epsilon_i\in e^P_i$ and $w^\epsilon_j\in e^H_j$.

\subsubsection{Lexical Similarity Matrix  ($\mathbf{M}_l$)}
\label{subsubsec:matriz_lexica}
To capture word-level relationships, we construct a similarity matrix between all lemmas in $P$ and $H$ (including attributes). We define this matrix $\mathbf{M}_l \in [0,1]^{p \times q}$ analogously:
\[
(\mathbf{M}_l)_{ij} = \sigma(\mathbf{v}(w^P_i), \mathbf{v}(w^H_j)),\;0\leq i\leq p\;\text{ and }0\leq j\leq q,
\]
where $w^P_i\in \mathcal{L}_P$,  $w^H_j\in \mathcal{L}_H$, and $p = |\mathcal{L}_P|$, $q = |\mathcal{L}_H|$.

\begin{figure}[h!]
    \centering
    \begin{subfigure}[b]{\textwidth}
    \centering
    \includegraphics[width=0.75\textwidth]{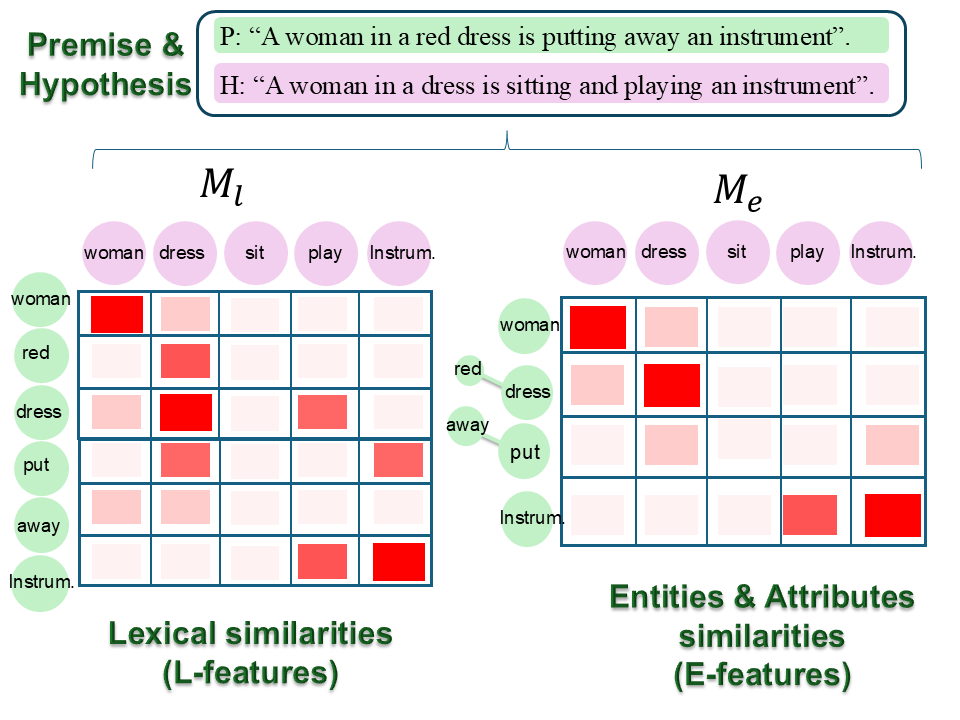}
    \caption{}\label{fig:pipeline_a}
    \end{subfigure}
    \begin{subfigure}[b]{\textwidth}
    \centering
    \includegraphics[width=0.65\textwidth]{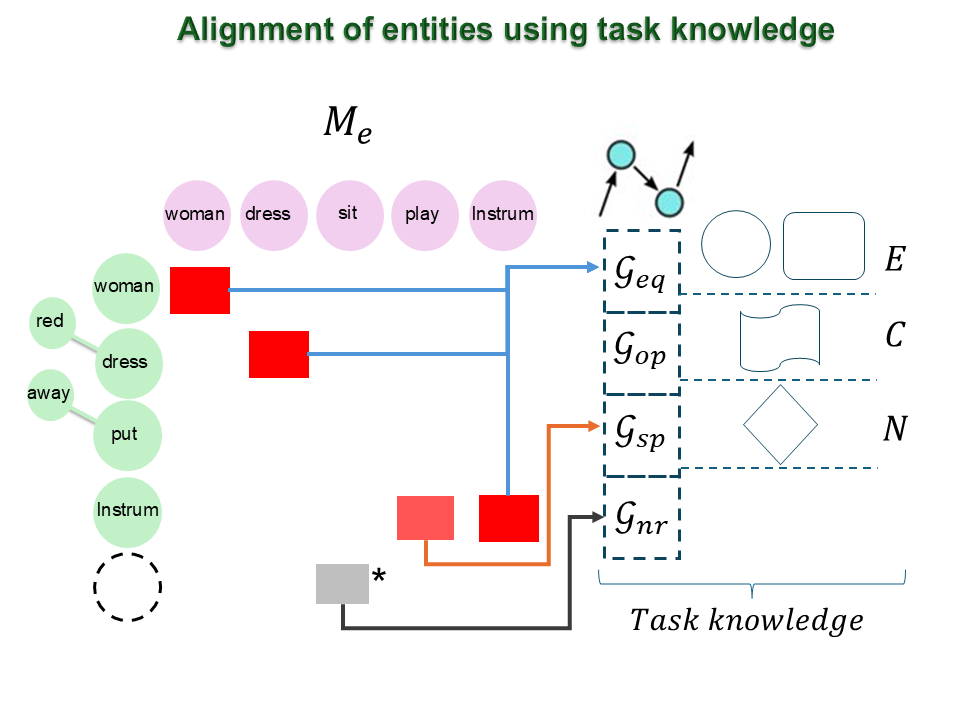}
    \caption{}\label{fig:pipeline_b}
    \end{subfigure}
    \caption{Proposed process for feature generation and extraction. a) Text processing and b) Identification of semantic categories.
    The starred gray box illustrates the case where some entity of $H$ has no matching entity in $P$.
    }
    \label{fig:pipeline}
\end{figure}

Both matrices preserve the directionality $P \rightarrow H$, in that the rows of $M_e \text{(or }M_l\text{)}$ are indexed by entities $e^P_i\in \mathcal{E}_P$ (or lemmatized words $w_i^P\in\mathcal{L}_P$), and their respective columns are indexed by the entities $e^H_j\in\mathcal{E}_H$ (or lemmatized words $w_j^H\in\mathcal{L}_H$)---see Fig. \ref{fig:pipeline_a}.

\subsection{Semantic categorization of relationships}
\label{subsec:categorizacion}

In what follows, we treat each word in a sentence as a \textit{concept} in the
sense of ConceptNet: a lemmatized lexical item that may participate in directed relational edges within the knowledge graph.

We follow the methodology of \citep{TORRESMORENO2026114825}. Consider two concepts $c_1$ and $c_2$ (note that the notion of concept applies to each member of the entities and their respective attributes). Hierarchical levels in ConceptNet are interpreted in terms of \textit{extensional generality}: a concept $c_2$ is at
a higher level than $c_1$ if the extension of $c_2$ properly includes that of $c_1$ (e.g., \textit{animal} $\supset$ \textit{dog}), corresponding to a hypernymy or $IsA$ relation directed from $c_1$ upward to $c_2$. This interpretation is consistent with the distributional entropy account of \citet{santus2014}, where more general
concepts occupy more diverse contexts.
 We define semantic compatibility and incompatibility between $c_1$ and $c_2$ in a general setting as follows.
\begin{definition}[Semantic Compatibility, SC]
\label{def:sc}
We say that $c_2$ has a relation of Semantic Compatibility with $c_1$---written $\mathrm{SC}(c_1, c_2)=1$---if there exists a directed path from $c_1$ to $c_2$ in the conceptual graph such that $c_2$ is at a strictly higher level of extensional generality
(\textbf{generalization relation}, $\mathcal{R}_{gn}$) or $c_2$ is at the same level with \textbf{equivalent} meaning (\textbf{equivalence relation}, $\mathcal{R}_{eq}$).
Formally:
\begin{equation}
\mathrm{SC}(c_1, c_2) = 1 \iff \mathcal{R}_{gn}(c_1, c_2, l) = 1 \;\lor\; \mathcal{R}_{eq}(c_1, c_2) = 1
\end{equation}
where $l \geq 1$ denotes the number of hierarchical steps separating $c_1$ and $c_2$
in the \emph{upward} direction.
\end{definition}

\begin{definition}[Semantic Incompatibility, SI]
\label{def:si}
We say that $c_2$ has a relation of \textbf{Semantic Incompatibility} with $c_1$---written $\mathrm{SI}(c_1, c_2) = 1$---if there exists a directed path from $c_1$ to $c_2$ such that $c_2$ is at a strictly lower level of extensional generality (specificity relation, $\mathcal{R}_{sp}$) or at the same level with a contrastive meaning
(opposition relation, $\mathcal{R}_{op}$). Formally:
\begin{equation}
\mathrm{SI}(c_1, c_2) = 1 \iff \mathcal{R}_{sp}(c_1, c_2, l) = 1 \;\lor\; \mathcal{R}_{op}(c_1, c_2) = 1
\end{equation}
where $l \geq 1$ denotes the number of hierarchical steps separating $c_1$ and $c_2$ in the \emph{dowward} direction.
\end{definition}
Given two entities $e^H_j\in\mathcal{E}_H$ and $e^P_i\in\mathcal{E}_P$, we say that any semantic relation $\mathbb{SR}$
of those defined above holds for $(e^P_j,e^H_j)$, or $\mathcal{R}_{\mathbb{SR}}(e^P_j,e^H_j)=1$, if the following expression evaluates to $1$:
\begin{equation}
\label{eq:Rsr}
    \left[\mathcal{R}_{\mathbb{SR}}(w_i^{\epsilon_P},w_j^{\epsilon_H})=1\right] \wedge \left[\forall k, \forall r \; \mathcal{R}_{\mathbb{SR}}(w_k^{a_P},w_r^{a_H})=1\right]
\end{equation}
Equation~(\ref{eq:Rsr}) employs universal quantification over the attribute sets
$\{w^{aP}_k\}$ and $\{w^{aH}_r\}$. Hence, we adopt the following convention. When one or both entities carry no attributes, the universal quantification is satisfied vacuously, and the semantic relation between the entities is determined solely by the head concept. Indeed, an entity without modifiers imposes no additional constraints on its referent, so the absence of attributes should not block an otherwise valid semantic relation between heads. For example, if $P$ contains \textit{car} and $H$ contains \textit{vehicle} (both without attributes), the generalization relation between the heads is sufficient to establish $\mathcal{R}_{gn}$, and the empty attribute condition does not interfere. Conversely, when attributes are present on both sides, \textit{all} attribute pairs must satisfy the relation, ensuring that entity-level compatibility is not granted by head agreement alone when modifiers introduce contradictory
or more specific information---as in the contrast between \textit{red car} and \textit{blue car}, where the opposition between color attributes correctly blocks the compatibility relation despite the head equivalence. Thus, each entity $e^H_j\in\mathcal{E}_H$ is paired with some entity $e^P_i\in\mathcal{E}_P$ using ConceptNet relations and verifying attribute compatibility.

We now define categorical sets $\mathcal{G}_{\text{eq}}$, $\mathcal{G}_{\text{op}}$, $\mathcal{G}_{\text{sp}}$, $\mathcal{G}_{\text{nr}}$  that correspond respectively to the semantic categories of Equivalence/Generalization, Opposition, Specificity, and No Relation (see Fig. \ref{fig:pipeline_b}), such that $\{(e^P_j,e^H_j)\}\subset\mathcal{E}_P\times \mathcal{E}_H$.
Notice that No Relation means that some entity in $H$ has no matching entity in $P$ by ConceptNet.
$$
\begin{array}{ccl}
     \mathcal{G}_{eq} &:=& \{(e_i^P,e_j^H) |\,  \mathcal{R}_{gn}(e_i^P,e_j^H)=1 \text{ or } \mathcal{R}_{eq}(e_i^P,e_j^H)=1
     \},  \\
     \mathcal{G}_{op} &:=& \{(e_i^P,e_j^H) |\, \mathcal{R}_{op}(e_i^P,e_j^H)=1 \},\\
\mathcal{G}_{sp} &:=& \{(e_i^P,e_j^H) |\, \mathcal{R}_{sp}(e_i^P,e_j^H)=1 \},\\
\mathcal{G}_{nr} &:=& \{(e_j^H,) |\, e_j^H\text{ has no relation in ConceptNet with any }e_i^P \in \mathcal{E}_P \}.
\end{array}
$$
The rationale of the categories respects the directionality $P \rightarrow H$; thus, we can say whether the entities in $H$ are equivalent or more general or more specific or opposite with respect to entities in $P$ according to their categorization. Moreover, following \citep{TORRESMORENO2026114825}, we align these semantic categories to the task classes as follows (Fig. \ref{fig:pipeline_b}):
$$
\begin{array}{cl}
     \mathcal{G}_{eq} &\text{ aligns with }\longrightarrow Entailment  \\
     \mathcal{G}_{op} &\text{ aligns with }\longrightarrow Contradiction  \\
     \mathcal{G}_{sp} &\text{ aligns with }\longrightarrow Neutral  \\
     \mathcal{G}_{nr} &\text{ has no particular alignment }
\end{array}
$$

Our semantic categorization and alignments are grounded in well-established work in lexical and formal semantics. Classical accounts distinguish relations such as synonymy, hyponymy, and antonymy, among others \citep{lyons1977semantics,cruse2004meaning,murphy2003semantic}, which can be naturally interpreted as set-theoretic relations between denotations. Under this view, for example, synonymy corresponds to equivalence, hyponymy to set inclusion, and antonymy to various forms of semantic exclusion. These relations have long been known to connect with propositional relations, specifically entailment \citep{cruse2004meaning,hurford2007semantics,jeffries1998meaning}. In fact, entailment relations between propositions have traditionally been a useful tool to define sense relations for semanticists \citep{lyons1977semantics}. Based on these well-established theoretical connections between lexical-semantic relations and proposition entailments, our rationale is justified as follows:

\begin{itemize}
\item General and equivalent relations correspond with Entailment: both the relations included under “general relations” (such as hyperonymy, meronymy….) and those under “equivalent relations” (synonymy), impose a relation between sentences such that the truth of P (e.g., there is a man) makes H necessarily true (e.g. there is a person) \citep{cruse2004meaning,hurford2007semantics}.
\item Opposite and difference relations, which encode semantic incompatibility, correspond with Contradiction. First, opposite and difference relations (antonyms, complementaries, converses, directional opposites, as well as co-hyponyms and co-meronyms etc.) include lexical items that cannot be predicated of the same entity as true simultaneously. These lexical relations impose a logical contrary relation between sentences: the truth of one entails the falsity of the other \citep{cruse2004meaning,jeffries1998meaning}.  Consequently, if P and H differ in a pair of concepts that are opposites or different, P does not entail H: The truth of P does not allow to infer the truth of H. Second, regarding “concrete relations”, the truth of the more general term (higher in the lexical hierarchy), by definition, does not entail the truth of the less general term subordinated in the scale \citep{cruse2004meaning,jeffries1998meaning}.
\item Relations of specificity do not license entailment but instead give rise to neutrality. When the hypothesis is more specific than the premise, it introduces additional information that is not guaranteed by the premise and therefore cannot be inferred from it. In set-theoretic terms, this corresponds to a narrowing of the denotation: while the extension of the hypothesis may be included within that of the premise, the direction required for entailment is reversed, and thus the inference does not hold. This asymmetry is well documented in formal semantics. Such cases fall under the category of neutral, since the hypothesis may be true given the premise, but is not necessarily so.

Furthermore, neutrality in NLI is a heterogeneous and theoretically
contested category. Unlike entailment and contradiction, which correspond
to well-defined logical relations with relatively stable annotation
criteria, the neutral class has been a persistent source of disagreement
in the literature: it groups together pairs that are thematically
independent, pairs where the inference is blocked by pragmatic rather
than semantic factors, and pairs where $H$ introduces information more
specific than $P$---all under a single label that conflates logically
distinct phenomena \citep{kalouli-etal-2023-curing, gururangan-etal-2018-annotation}. This annotation
instability is not incidental: \citet{kalouli-etal-2023-curing} show that inter-annotator agreement is substantially lower for neutral pairs than for entailment and contradiction across multiple NLI datasets, and that models trained on neutral examples are more susceptible to learning spurious correlations precisely because the category itself is less coherent.
Our framework addresses this heterogeneity operationally: the categories $\mathcal{G}_{sp}$ and $\mathcal{G}_{nr}$ provide distinct operational handles on two of the most
common sources of neutrality---specificity and absence of
relation---while the classifier is exposed to their interaction with equivalence and opposition signals  simultaneously. This explicit decomposition allows the model to distinguish, at least partially, between types of neutral pairs that a single label obscures, and provides a principled basis for diagnosing the cases where neutral classification fails.
\end{itemize}

In this sense, the alignment between semantic categories and task classes should be regarded as a necessary, but not sufficient, condition. For a pair to be classified as entailment, all relations between entities should ideally correspond to relations of generality or equivalence ($\mathcal{G}_{eq}$), without the presence of opposition or specificity. However, the existence of relations of generality or equivalence does not necessarily imply entailment, since such relations may also occur in pairs belonging to other classes. In contradiction pairs, several relations of generality and equivalence may coexist with a single relation of opposition ($\mathcal{G}_{op}$). Nevertheless, it is precisely this opposition relation, regardless of its relative frequency, that determines the class and overrides any positive evidence. Similarly, in the neutral class, relations of generality, equivalence, and even moderate opposition may occur simultaneously. What distinguishes this class is the presence of specificity ($\mathcal{G}_{sp}$) or unrelated new information ($\mathcal{G}_{nr}$), which prevents both entailment and direct contradiction.

This asymmetry---where opposition is determinative, specificity and no-relations introduce ambiguity, and generality or equivalence require contextual interpretation---supports the need to model not only the presence of each relation type, but also their interactions and relative contributions. As noted above, the presence of equivalence or generality relations does not guarantee entailment, nor does their absence necessarily imply contradiction. Instead, class membership depends on the interaction between these positive signals and the presence of opposition or specificity. By explicitly modeling these interactions, the proposed feature set can disentangle the often ambiguous contribution of generality and equivalence, thereby improving class discrimination.

Furthermore, working with knowledge bases and semantic searches requires establishing priorities among candidate relations. Such prioritization may exclude relations whose information could still be relevant for decision-making. Conversely, identifying a general relation first may obscure a more informative specific relation; in this sense, generalization may introduce noise into the decision process. This situation may arise across the different types of semantic relations considered in this work. The proposed set of measures seeks to mitigate these effects by focusing the analysis on combinations of specific relations, thereby reducing the influence of noise.

Thus, by measuring the contribution of each semantic group both collectively and individually, it becomes possible to capture the distinctive characteristics associated with each class. The following section presents this proposal in detail.

\section{The SLITE Model: Structural-Relational and
Distributional-Informational Features}
\label{sec:caracteristicas}

We refer to our model as \textbf{SLITE}
(\textbf{S}emantic \textbf{L}ayers for
\textbf{I}nterpretable \textbf{T}extual
\textbf{E}ntailment), a name that reflects
its core design principle: the integration
of two complementary layers of semantic
analysis, Structural-Relational and
Distributional-Informational, into a unified,
transparent representation of the relationship
between premise and hypothesis.

We propose 17 features organized across these
two layers:
\begin{itemize}
    \item \textbf{Entity level:} 3
    Structural-Relational features (E1, E2, E8)
    and 5 Distributional-Informational features
    (E3--E7).
    \item \textbf{Lexical level:} 5
    Structural-Relational features (L13--L17)
    and 4 Distributional-Informational features
    (L9--L12).
\end{itemize}

\subsection{Entity-Level Features: Structural-Relational
    and Distributional-Informational}
\label{subsubsec:prop_sem}
Entity-based features are at the core of our approach. The construction of entities allows us to model the compositional nature of language in order to give rise to semantically richer lexical units. In this way, the search for semantic relationships between these composite entities allows us to identify non-superficial relationships. For example, if $H$ contains ‘\textit{red clothes}’ and $P$ contains ‘\textit{red car}’, at the lexical level (isolated words) we might find that ‘\textit{car}’-‘\textit{clothes}’ have no relationship but ‘\textit{red}’-‘\textit{red}’ do, which could lead to ambiguity in determining whether $P\to H$. Using entities, we would find that ‘\textit{red car}’-‘\textit{red clothes}’ are two different things, even though their attributes are similar, so there is no ambiguity in saying that $P\to H$ does not hold. Below, we formalize the process to extract our features.

We start by defining the following measure:
$$
     \text{$\rho_{R}$}  = \frac{|\mathcal{G}_{\text{$R$}}|}{n}, \label{eq:prop_equiv}
$$
where $R$ $\in$ \{\text{eq}, \text{op}, \text{sp}, \text{nr}\}. The intuition behind this ratio is to quantify the proportion of entities in $H$ that have relations belonging to any of the semantic categories. For instance, the ratio $\rho_{eq}$ quantifies the proportion of pairs $(e^P_i,e^H_j)\in \mathcal{G}_{eq}$, such that entities in $H$ are equivalent to, or more general, than their counterparts in $P$. We might think about this ratio as the weight of equivalence or generalization relations in $H$. Notice that $$
\rho_{eq}+\rho_{op}+\rho_{sp}+\rho_{nr} = 1,
$$
which means that the sets $\mathcal{G}_R$ are mutually exclusive, $R \in \{eq,op,sp,nr\}$ ---i.e. any entity of $H$ belongs to a single set.

We now introduce some notation. First, notice that in RTE the focus is on directionality: the meaning of $H$ must be contained within the meaning of $P$ for there to be entailment. Therefore, if we remove elements from $H$, the remaining elements must preserve the meaning of $P\to H$ if entailment holds. The assumption is that this degree of meaning preservation can be measured using the remaining similarity values; we say that there is a degree of \emph{alignment}.

Let $\mathbf{M}_e^{(-\mathcal{G}^H_{R})}$ be the submatrix resulting from \emph{removing the columns} of $\mathbf{M}_e$ corresponding to the entities $e^H_j\in \mathcal{E}_{H}$ that belong to $\mathcal{G}_R$.
We will write $\lvert\mathbf{M}_e^{(-\mathcal{G}^H_{\text{R}})}\rvert$ to represent the number of elements of $\mathbf{M}_e^{(-\mathcal{G}^H_{\text{R}})}$. Thus, $\mathbf{M}_e^{(-\mathcal{G}^H_{R})}$ contains the cosine similarity values between the embeddings of all the entities of $P$ and only those entities of $H$ that do not belong to the relations $\mathcal{G}_R$. For example, Fig. \ref{fig:pipeline_b} shows that three entities of $H$ belong to $\mathcal{G}_{eq}$: \textit{woman, dress} and \textit{instrument}. Therefore, since $\lvert\mathcal{G}_{eq}\rvert=3$, we have that while $\mathbf{M}_e\in \mathbb{R}^{4\times 5}$, $\mathbf{M}_e^{(-\mathcal{G}^H_{eq})}\in \mathbb{R}^{4\times 2}$, where the columns corresponding to the three entities of $H$ were removed, and $\lvert\mathbf{M}_e^{(-\mathcal{G}^H_{\text{eq}})}\rvert=8$.
Similarly, we define $\mathbf{M}_e^{(-\mathcal{G}^P_{R})}$ to be the matrix resulting from \emph{removing the rows} of $\mathbf{M}_e$ corresponding to the entities $e^P_i\in \mathcal{G}_{R}$. Continuing with the same example, $\mathbf{M}_e^{(-\mathcal{G}^P_{eq})}\in \mathbb{R}^{1\times 5}$, where the rows corresponding to the three matching entities of $P$ were removed, and $\lvert\mathbf{M}_e^{(-\mathcal{G}^P_{\text{eq}})}\rvert=5$.

On the other hand, if we remove all information about entities linked by a relation $R$, we are reducing the scope of the meaning of $P$ and $H$, setting aside elements that were previously linked by $R$, and exploring the preservation of meaning through the remaining similarity values.  Thus, we will write $\mathbf{M}_e^{(-\mathcal{G}_{R})}$ to define the submatrix resulting from \emph{removing all the rows and columns} that correspond to pairs $(e^P_i,e^H_j)\in \mathcal{G}_{R}$. Continuing with the example above,  $\mathbf{M}_e^{(-\mathcal{G}_{eq})}\in \mathbb{R}^{1\times 2}$ and $\lvert\mathbf{M}_e^{(-\mathcal{G}_{\text{eq}})}\rvert=2$.

Intuitively, these submatrices contain information about the interactions between the distinct groups of relations that remain.
In the sequel, we are interested in computing the Shannon entropy $\textbf{H}$ of this kind of matrices. The entropy of a distribution $X$ is defined as $\textbf{H}(X)=-\sum_{x\in X} p(x)\log p(x)$. Thus,
$$
\textbf{ H}\left(\mathbf{M}_e^{(-\mathcal{G}^H_{R})}\right) := -\sum_{x\in \mathbf{M}_e^{(-\mathcal{G}^H_{R})}} x\log x
$$
 $\textbf{  H}\left(\mathbf{M}_e^{(-\mathcal{G}^H_{R})}\right)$ estimates the variability of the cosine similarities that remain in $\mathbf{M}_e$ after removing the corresponding columns verifying the relation $R$.
The features of the entity level are listed below.

\begin{itemize}
    \item[\textbf{$E1:$}] \textbf{Strength of vertical relations and equivalence:} Proportion of entities in $H$ having  equivalence or general and specificity relations.
    \[
        \textbf{$E1:$} = \rho_{eq}+\rho_{sp}=\frac{|\mathcal{G}_{eq}| + |\mathcal{G}_{sp}|}{n}
    \]
    \item[\textbf{$E2:$}] \textbf{Strength of generality, equivalence and ambiguity:} Proportion of entities in $H$ with equivalence or generality and no relation to entities of P.
    \[
        \textbf{$E2:$} = \rho_{eq}+\rho_{nr}= \frac{|\mathcal{G}_{eq}| + |\mathcal{G}_{\text{${nr}$}}|}{n}
    \]
    \item[\textbf{$E3:$}] \textbf{Alignment strength without opposition:} Average similarity value in the matrix excluding entities $e^H_j\in \mathcal{G}_{op}$. The intuition behind this measure is that removing the opposite relations should lead to average similarity values consistent with entailment values.
    \[
    \textbf{$E3:$} = \frac{1}{\lvert\mathbf{M}_e^{(-\mathcal{G}^H_{\text{op}})}\rvert} \sum_{i,j} (\mathbf{M}_e^{(-\mathcal{G}^H_{\text{op}})})_{ij}
    \]

    \item[\textbf{$E4:$}] \textbf{Information loss without equivalence and opposition alignments:} Entropy difference between the complete similarity matrix and the corresponding submatrix without columns of equivalent and opposition entities.
    \[
    \textbf{$E4:$}= \textbf{H}(\mathbf{M}_e) - \textbf{ H}\left(\mathbf{M}_e^{(-(\mathcal{G}^H_{\text{eq}} \,\cup\, \mathcal{G}^H_{\text{op}})}\right)
    \]

    \item[\textbf{$E5:$}] \textbf{Average maximum alignment without equivalence and generality:} Proportion of the maximum similarities per column for non-equivalent and non-general entities. The intuition is that if ConceptNet's general or equivalent relationships fail to capture all possible relationships between entities in $P$ and $H$, this measure would capture their average maximum similarity in the embedding space.
    \[
    \textbf{$E5:$}= \frac{1}{j} \sum_{j} \max_{1 \leq i \leq m} (\mathbf{M}_e^{(-\mathcal{G}^H_{\text{eq}})})_{ij}
    \]

    \item[\textbf{$E6:$}] \textbf{Entropy of the alignment without generality and equivalence:} Measures the uncertainty when equivalent and general entities of $H$ are excluded.

    \[
    \textbf{$E6:$}= \textbf{H} \left(\mathbf{M}_e^{(-\mathcal{G}^H_{\text{eq}})}\right)
    \]

    \item[\textbf{$E7:$}] \textbf{Entropy of the submatrix without specificity:} Quantifies the information when removing entities $ (e^P_i,e^H_j)\in\mathcal{G}_{sp}$.
    \[
    \textbf{$E7:$}= \textbf{ H}\left(\mathbf{M}_e^{(-\mathcal{G}_{\text{sp}})}\right)
    \]

\end{itemize}

\begin{itemize}
    \item[\textbf{$E8:$}] \textbf{Weighting of group relevance:} We define a base entailment score that synthesizes categorical information supported by the entropy of missing entities:
 \[
\textbf{$E8:$} =
\begin{cases}
\dfrac{\rho_{eq} + 0.1 \rho_{sp}}{1 + \rho_{nr} * \textbf{H}\left(\mathbf{M}_e^{(-\mathcal{G}_{\text{nr}})}\right)}, & \text{if } |\mathcal{G}^H_{\text{op}}| = 0 \\
0.0, & \text{if } |\mathcal{G}_{\text{op}}| > 0
\end{cases}
\]
\end{itemize}

It weights the proportions of the groups of relationships identified between $P$ and $H$. The intuition is that if only generality relationships exist, the other groups would be empty, so the measure takes the value 1. If there is at least one opposition relation, this measure returns the value 0. Therefore, the rationale behind this measure is to return a value between 0 and 1 by weighting the relative value of the other relations, namely specificity relations and unidentified relations. The former contribute to increasing the score, whilst the latter decrease the score based on the entropy of the corresponding submatrix without that group.

The coefficients in $E_8$ were determined through a grid search on the development set, independently of the test partition used for evaluation. Specifically, the weight $0.1$ assigned to $\rho_{sp}$ reflects the empirical finding that specificity relations contribute a secondary positive signal for entailment, attenuated relative to equivalence relations by an order of
magnitude. The entropy-based penalty for $\rho_{nr}$ was selected over a fixed penalty because it scales with the
informational content of the unmatched entities, providing a more nuanced account of cases where no relation is found.

Together, these eight features provide a multidimensional profile of the relationship between premise and hypothesis at the entity level, combining quantitative rigor with semantic sensitivity. This representation allows the model to learn not only what types of relationships exist, but also how much information they carry and how they are statistically structured.

\subsection{Lexical-Level Features: Distributional-Informational}
\label{subsubsec:prop_lex}

Lexical features complement entity-based analysis operating at the individual word level, capturing both surface similarities and distributional properties of embeddings. This finer level of granularity allows the model to access alignment signals that transcend entity structure, including relationships between function words, modifiers, and elements that do not constitute main entities. Derived from the identification of $\mathcal{G}_{\text{eq}}$, $\mathcal{G}_{\text{op}}$, $\mathcal{G}_{\text{sp}}$, $\mathcal{G}_{\text{nr}}$ relationships, we retrieve all the words that make up the whole entities to obtain new sets $\mathcal{W}_{\text{eq}}$, $\mathcal{W}_{\text{op}}$, $\mathcal{W}_{\text{sp}}$, $\mathcal{W}_{\text{nr}}$ as follows.
$$
\begin{array}{ccl}
     \mathcal{W}_{eq} &=& \{w_j^H \,|\, w_j^H \in e_j^H,\; \forall e_j^H \in \mathcal{G}_{eq}\}\\
     \mathcal{W}_{op} &=& \{w_j^H \,|\, w_j^H \in e_j^H,\; \forall e_j^H \in \mathcal{G}_{op}\}\\
     \mathcal{W}_{sp} &=& \{w_j^H \,|\, w_j^H \in e_j^H,\; \forall e_j^H \in \mathcal{G}_{sp}\}\\
     \mathcal{W}_{nr} &=& \{w_j^H \,|\, w_j^H \in e_j^H,\; \forall e_j^H \in \mathcal{G}_{nr}\}.
\end{array}
$$

For example, if the entity ‘red dress’ in $H$ is classified as equivalent ($\mathcal{G}_{\text{eq}}$) to an entity in $P$, then the words ‘red’ and ‘dress’ are all added to $\mathcal{W}_{\text{eq}}$. This word-level representation allows us to capture the lexical contribution of each semantic category beyond the entity level, enabling finer-grained analysis of information flow, lexical alignment, and contradiction signals. The same principle applies to the other categories: $\mathcal{W}_{\text{op}}$ contains words from opposition entities, $\mathcal{W}_{\text{sp}}$ from specificity entities, and $\mathcal{W}_{\text{nr}}$ from entities with no relation to $P$.

For the lexical matrix $\mathbf{M}_l \in [0,1]^{p \times q}$, let $\mathbf{M}_l^{(-\mathcal{W}_{R})}$ be the submatrix resulting from \emph{removing the columns} of $\mathbf{M}_l$ corresponding to the words $w^H_j\in \mathcal{W}_R$.
We will write $\lvert\mathbf{M}_e^{(-\mathcal{W}_{\text{R}})}\rvert$ to represent the number of elements of $\mathbf{M}_l^{(-\mathcal{W}_{\text{R}})}$. Thus, $\mathbf{M}_l^{(-\mathcal{W}_{R})}$ contains the cosine similarity values between the embeddings of all the words of $P$ and only those words of $H$ that do not belong to the relations $\mathcal{W}_R$. We define analogous operations and the generated features:

\begin{itemize}
    \item[$L9:$] \textbf{Average maximum alignment:} Sum of the maximum similarities per column, normalized by the number of words in $H$.
    \[
    \textbf{$L9:$} = \frac{1}{q} \sum_{j=1}^{q} \max_{1 \leq i \leq p} (\mathbf{M}_l)_{ij}
    \]

    \item[$L10:$] \textbf{Average maximum alignment without equivalent and specificity terms:} Maximum information conveyed by opposition or unrelated words.
    \[
    \textbf{$L10:$} = \frac{1}{q} \sum_{j=1}^q \max_{1 \leq i \leq p}(\mathbf{M}_l^{-(\mathcal{W}_{\text{eq}} \cup \mathcal{W}_{\text{sp}})})_{ij}
    \]
    \item[$L11:$] \textbf{Average alignment without specificity and unrelated terms:} Average value for words without semantic relation of specificity and unrelated.
    \[
    \textbf{$L11:$} = \frac{1}{|\mathbf{M}_l^{-(\mathcal{W}_{\text{sp}} \cup \mathcal{W}_{\text{nr}})}|} \sum_{i,j} (\mathbf{M}_l^{-(\mathcal{W}_{\text{sp}} \cup \mathcal{W}_{\text{nr}})})_{ij}
    \]

    \item[$L12:$] \textbf{Specificity Lexical Entropy Transfer:}
Measures the directional flow of information from the premise to the hypothesis, without considering the terms (entities with attributes) in the hypothesis classified as opposites ($\mathcal{W}_{\text{op}}$) or unrelated ($\mathcal{W}_{\text{nr}}$). We calculate entropy transfer as \citep{PhysRevLett.85.461,KAISER200243}:

\[
\textbf{$L12:$} = T_{X \to Y}^{\text{(safe)}} \quad \text{with} \quad
\begin{cases}
X = \mathbf{M}_l^{-(\mathcal{W}_{\text{eq}} \cup \mathcal{W}_{\text{sp}})} \\
Y = \left(\mathbf{M}_l^{-(\mathcal{W}_{\text{eq}} \cup \mathcal{W}_{\text{sp}})}\right)^\top \\
delay = 1
\end{cases}
\]
where $\mathbf{M}_l^{(-\mathcal{W}_{\text{eq}} \cup \mathcal{W}_{\text{sp}})}$ denotes the lexical matrix $\mathbf{M}_l$ after removing the columns corresponding to equivalent words ($\mathcal{W}_{\text{eq}}$) and specific words ($\mathcal{W}_ {\text{sp}}$) and where $T_{X \to Y}^{\text{(safe)}}$ implements the robust entropy transfer algorithm with length validation and sequence synchronization.
\end{itemize}

\subsection{Lexical-Level Features: Structural-Relational}
\label{subsubsec:str_sim}
We complete the proposal with a set of symbolic semantic similarity features. To this end, the Jaro algorithm \citep{Jaro1989} was implemented and extended to sentences at the lexical-semantic level with ConceptNet relations. Our proposal implements a significantly modified version of the Jaro distance, specifically adapted for RTE tasks. While preserving the concept of a search window based on the maximum sequence length, the count of transpositions, and the classic three-component formula, the proposed similarity measure works with words rather than characters, and the matching criterion is not based on exact equality but on semantic relationships as given by Definitions \ref{def:sc} and \ref{def:si}.
Furthermore, it incorporates a polarity analysis that considers negation words to distinguish between affirmations and negations.

Central to the proposed measure is the notion of match. Its definition is built upon three interconnected components that collectively determine when two words can be considered a valid match, as formalized in Definitions \ref{def:pol} through \ref{def:ska_inv_match} below.

\begin{definition}[Polarity match]
\label{def:pol}
Given a sequence of words $X = (x_1, \ldots, x_k)$ and a set of negator words $\mathcal{N}$, we define the \textit{polarity} of a word $x_i$ as:
\[
\text{pol}(x_i) = \begin{cases}
1 & \text{if } i = 1 \text{ or } x_{i-1} \notin \mathcal{N} \\
0 & \text{otherwise}
\end{cases}
\]

where $1$ represents an affirmative state and $0$ represents a negated state.

We say that two words $x_i$ and $x_j$ have a \emph{polarity match}, if pol($x_i$)$=$pol($x_j$). In this case, we write $\mathcal{P}(x_i,x_j)=1$, otherwise $\mathcal{P}(x_i,x_j)=0$.
\end{definition}

\begin{definition}[$\mathbb{SR}$ valid match score.]
\label {def:ska_match}
A pair $(p_i, h_j)$ constitutes a \textit{match} if the following expression evaluates to $1$:
\vspace{1em}

\scalebox{0.9}{
\[
m_{(i,j),[l\geq 1]}=\Big[\mathcal{P}(p_i, h_j) \;\wedge\; \big(\mathcal{R}_{gn}(p_i, h_j,l) \vee \mathcal{R}_{eq}(p_i, h_j) \big)\Big] \vee \Big[ \neg \mathcal{P}(p_i, h_j) \;\wedge\; \mathcal{R}_{op}(p_i, h_j)\Big],
\]
}
\vspace{1em}

In this expression, the parameter $l$ represents the distance from the hierarchical level (upward or downward) where the relationship between $p_i$ and $h_j$ may exist in the knowledge base.
\vspace{1em}

Additionally, for a pair $(p_i, h_j)$ to be considered for matching, it must satisfy the distance constraint:
$\displaystyle |i - j| \leq \left\lfloor \frac{\max(|P|, |H|)}{2} \right\rfloor$.
\\
\\
Thus, we define the $\mathbb{SR}$ \emph{match} score $m_{\mathbb{SR}}$ between the premise $P$ and hypothesis $H$ as:
$$
m_{\mathbb{SR}}(P,H,l\geq 1) = \sum_{\forall i,j} m_{(i,j),[l\geq 1]}=1.
$$
We define a semantic mismatch coefficient $\tau$ as:
$$\tau = \frac{1}{2} \sum_{\forall i,j} m_{(i,j),[l\geq 1]}=0 $$
\end{definition}

The condition $l \geq 1$ represents the possibility that the semantic relationship can occur at an arbitrarily distant level in the hierarchy. In other words, if $l=1$, it means that the relationship is found at the immediately adjacent level (up or down). When we look for semantic relationships at distant hierarchical levels, we are making use of the property of transitivity, which allows us to establish non-immediate links between concepts.

Definition \ref{def:ska_match}, inherited from the original Jaro metric, restricts matching to words that are reasonably close in the sequential order of the texts, preserving the notion of local structure. We now propose a new similarity measure \emph{Jaro RTE with} $\mathbb{SR}$, which calculates the directional similarity between two sentences $P$ and $H$ based on valid matches $m_{\mathbb{SR}}$, where we have dropped the explicit dependence on the parameters for readability purposes. The measure is defined as:
\begin{equation}
\label{jaro_torres}
    J_{\mathbb{SR}}(P,H,l\geq 1) =
\begin{cases}
0, & \text{if } m_{\mathbb{SR}} = 0,\\[6pt]
\dfrac{1}{3}\left( \dfrac{m_{\mathbb{SR}}}{|P|} + \dfrac{m_{\mathbb{SR}}}{|H|} + \dfrac{m_{\mathbb{SR}} - \tau}{m_{\mathbb{SR}}} \right), & \text{ otherwise.}
\end{cases}
\end{equation}

The similarity measure (\ref{jaro_torres}) captures relationships with semantic compatibility, which are consistent with the directionality of the task for the entailment class. Semantic incompatibility yields types of relationships that run counter to entailment. Therefore, we need to define the sufficient conditions for invalidating it.

\begin{definition}[$\mathbb{SR}$ invalid match]
\label{def:ska_inv_match}
A pair $(p_i, h_j)$ constitutes a \textit{contradictory match} if the following expression evaluates to $1$:
\vspace{1em}

\scalebox{0.9}{
\[
cm_{(i,j),[l\geq 1]} =\left[\mathcal{P}(p_i, h_j) \;\wedge\; \mathcal{R}_{op}(p_i, h_j) \right] \vee
\left[\neg \mathcal{P}(p_i, h_j) \;\wedge\; \big(\mathcal{R}_{gn}(p_i, h_j,l) \vee \mathcal{R}_{eq}(p_i, h_j) \big)\right]
\]
}
\vspace{1em}

We write $cm=1$ if there exists a contradictory match for any pair $(p_i, h_j)$ at any level. A pair $(p_i, h_j)$ constitutes a \textit{specificity match} if the following expression evaluates to $1$:
\[
sm_{(i,j),[l\geq 1]} = \mathcal{R}_{sp}(p_i,h_j,l)
\]

We write $sm=1$ if there exists a specificity match for any pair $(p_i, h_j)$ at any level.
\end{definition}

In the first case, \(cm\) captures contradictory situations. If there is an opposition relationship between a pair $(p_i,h_i)$ and its polarity is 1, this means that the two words are semantically incompatible. Also, if the pair is semantically compatible but its polarity is 0, this means that one contradicts the other.
Meanwhile, \(sm\) captures the situation of semantic incompatibility due to specificity.
This involves no logical contradiction, but rather a legitimate hierarchical alignment where \(h_j\) specifies or refines the meaning of \(p_i\).

The semantic links established at hierarchical levels immediately above the concepts of $P$ and $H$ are direct semantic relations; that is, there are no intermediate concepts linking them. In this sense, a valid match would give rise to an entailment. The extension of semantic links through concepts located at distant hierarchical levels is valid due to the transitivity of semantic compatibility relations, also suggesting an entailment. However, these distant links can lead to false positives when attempting to connect two concepts via words in the hierarchy that may be lexically ambiguous.
For this reason, we propose the following features, with different scope conditions in the semantic relation: $l\geq 1$ and $l=1$, and penalties for $cm=1$, $sm=1$, and absence of a match.

We now have the conceptual framework needed to propose the final set of lexical-semantic features.

\begin{itemize}
        \item[$L13:$] \textbf{Jaro RTE.}
        This feature focuses on all semantic relationships $P \rightarrow H$ ($l\geq 1$),
        with penalties $\alpha$ for \textit{contradictory match}, \textit{specificity match}, and no matching words in $H$.
 \[
J_{\text{RTE}}(P, H) = J_{\mathbb{SR}}(P, H, l\geq 1) \times \alpha
\]

where $\alpha$ is chosen according to the following precedence rules, where the penalty values were determined empirically by a grid search (only the highest penalty applies):

\[
\alpha =
\begin{cases}
0.01 & \text{if there exists any \textit{contradictory match} } (cm = 1) \\[4pt]
0.6  & \text{else if there exists any \textit{specificity match} } (sm = 1\text{ for any entity at any level}) \\[4pt]
0.8  & \text{else if there is significant no information in } H \ (1 - \frac{m_{\mathbb{SR}}}{|H|} > 0.1)\\[4pt]
1    & \text{otherwise (base case, entailment).}
\end{cases}
\]

\end{itemize}
This measure projects a value close to $1$ when there is consistency for the RTE class \textit{entailment}, taking into account equivalence and generality relations from $P$ to $H$. If an opposition relation is found, it penalizes the score such that it points to the \textit{contradiction} class with a value near $0$. The measure centers the value around $0.6$ when a specificity relation from $P$ to $H$ is found.
The term $1-\frac{m_{\mathbb{SR}}}{|H|}$ represents the proportion of words in $H$ that were not matched with any word in $P$, i.e., unpaired information. When this exceeds 10\%, the hypothesis introduces significant new information, suggesting a \textit{neutral} class.\\

\begin{itemize}
        \item[$L14:$]
        \textbf{Jaro RTE with opposition.} This feature focuses on semantic relationships with $l=1$ between $P \rightarrow H$,
        and penalties $\beta$ for \textit{contradictory match}.
        \[
J_{\text{RTE}_{op}}(P,H) = J_{\mathbb{SR}}(P,H,l=1) \times \beta
\]
\[
 \text{where }
\beta =
\begin{cases}
0.01 & \text{if there is a \textit{contradictory match} ($cm=1$)}\\
1 & \text{otherwise}
\end{cases}
\]

        \item[$L15:$] \textbf{Jaro RTE with specificity.} This feature focuses on semantic relationships with $l=1$ between $P \rightarrow H$ ,
        and penalties $\gamma$ for \textit{specificity match}.
        \[
J_{\text{RTE}_{sp}}(P,H) = J_{\mathbb{SR}}(P,H,l=1) \times \gamma
\]
\[
 \text{where }
\gamma =
\begin{cases}
0.01 & \text{if there is a \textit{specificity match} ($sm=1$)}\\
1 & \text{otherwise}
\end{cases}
\]
        \item[$L16:$] \textbf{Reverse Jaro RTE with opposition.} Reverse version of [$L14$] that focuses on semantic relationships with $l=1$ between $H \rightarrow P$ and penalties $\beta$ for \textit{contradictory match}. Useful for detecting asymmetries in the entailment relationship. Basically, this feature asks whether the Premise generalizes the Hypothesis, which is incompatible with the logic of RTE. It also provides context for interpreting the match ratio obtained by [$L17$].
        \[
J_{\text{RTE}_{op}}(H,P) = J_{\mathbb{SR}}(H,P,l=1) \times \beta
\]
    \item[$L17:$] \textbf{Lexical-Semantic match ratio.} This feature focuses on the proportion of matched words $m_{\mathbb{SR}}(H,P,l=1)$ with respect to the premise.

\[
LS_{\text{ratio}}(H,P) = \frac{m_{\mathbb{SR}}(H,P,l=1)}{|P|}
\]

This measure captures the coverage of the hypothesis by the premise, serving as a complementary indicator to the directional Jaro variants. It is particularly useful when combined with L16 to detect asymmetries in entailment: while L16 detects contradictions in the reverse direction, L17 quantifies how much of $P$ is actually supported by $H$.
\end{itemize}

\subsection{Feature Integration and Classification Model}
\label{subsec:clasificacion}

These measures capture complementary aspects of surface similarity enriched with semantic knowledge, providing a bridge between embedding-based analysis and traditional linguistic features. The features are organized into four groups as shown in Table \ref{tab:features_summary}.
\begin{table}[h]
\centering
\begin{tabular}{lcc}
\hline
\multirow{2}{*}{\textbf{Level}} & \multicolumn{2}{c}{\textbf{Layers}} \\ \cline{2-3}
& \textbf{Structural-Relational} & \textbf{Distributional-Informational} \\ \hline
\multirow{2}{*}{Entity}   \\
& \texttt{E1}, \texttt{E2}, \texttt{E8} & \texttt{E3} - \texttt{E7} \\ \hline
\multirow{2}{*}{Lexical} \\
&   \texttt{L13} - \texttt{L17}
&\texttt{L9} - \texttt{L12} \\ \hline

\end{tabular}
\caption{Distribution of features: Entities-SR (3) + Entities-DI (5) + Lexical-SR (5) + Lexical-DI (4)   = 17 features.}\label{tab:features_summary}
\end{table}

The 17 features are used as input for a Logistic Regression classifier. Logistic regression is selected for three primary reasons: first, its decision boundary is a linear combination of input features, allowing each coefficient to be directly interpreted as a measure of feature importance; second, it is computationally efficient during both training and inference; and third, it avoids introducing additional representational complexity that might obscure the linguistic analysis encoded within the features.

To ensure reproducibility and support future research, the SLITE implementation ---including feature extraction scripts and classification models--- is publicly available on GitHub\footnote{https://github.com/labsemco/SLITE}.

\section{Experiments and Results}
\label{sec:experimentos}

In this section, we detail the experimental setup, the datasets used, the comparison models, and present a comprehensive analysis of the performance of our proposal.

\subsection{Experimental Settings}
\label{subsec:configuracion}

The process applied was the one described in Section~\ref{sec:notacion}, which includes lemmatization and the construction of entity representations based on the dependency tree. Next, we calculate the features described in Section~\ref{sec:caracteristicas} in order to classify them using logistic regression and determine the class of the $P$ and $H$ pairs. The logistic regression classifier was implemented using the Scikit-learn library, performing five-fold cross-validation.

We evaluated our model on four widely used public benchmarks for the RTE task:

\begin{itemize}
    \item \textbf{SICK (Sentences Involving Compositional Knowledge)} \citep{marelli-etal-2014-sick}: This dataset contains 9,927 manually annotated sentence pairs with three relations: entailment, neutral, and contradiction. We use the official partition of training (4,500 pairs), development (500 pairs), and test (4,906 pairs).
    \item \textbf{SICK C-E (SICK Contradiction-Entailment)}: A variant of SICK where examples from the \textit{neutral} class have been removed, resulting in a binary classification problem between \textit{contradiction} and \textit{entailment} \citep{souza2025hybrid}. This version consists of 4,245 pairs in total, with 2,116 pairs in the test set. This partition allows for specific evaluation of the model's ability to distinguish between entailment and contradiction.
    \item \textbf{SICK correction for MonaLog}: A manually and rule-based corrected version of SICK proposed within the MonaLog framework \citep{hu-etal-2020-monalog}. This curation focuses on identifying and fixing specific labeling errors, including asymmetric annotations, and problematic word-level differences using WordNet-based rules. The authors manually reviewed 409 proposed changes and adjudicated disagreements to create a supplementary dataset with more precise labels.
    \item \textbf{SICK-LCS (SICK Logic and CommonSense)}: A re-annotated version of SICK proposal by \citep{kalouli-etal-2023-curing}. This disentanglement of inference types addresses inconsistencies and label noise present in the original SICK, allowing for separate evaluation of models on logical versus common-sense reasoning.
\end{itemize}

We compare our model with:
\begin{itemize}
    \item \textbf{IsoLex} \citep{souza2025hybrid}: A hybrid and explainable model for NLI that combines an Isolation Forest-based filter on BERT embeddings, lexical relation extraction using WordNet, and Word2Vec vector similarity. It is an ideal baseline due to its interpretable nature and specific evaluation in SICK C-E.
    \item \textbf{RoBERTa} \citep{liu2019roberta}: Reported by \cite{souza2025hybrid} as the state of the art for SICK C-E in their study. This model represents the upper limit of performance with approaches based on fine-tuned large language models (LLMs).
    \item \textbf{MonaLog} \citep{hu-etal-2020-monalog}: A lightweight natural language inference engine based on natural logic and monotonicity calculus. Unlike other logical systems, it operates directly on surface linguistic forms without translating into complex logical representations, using a small inventory of monotonicity facts about quantifiers and token polarity.
    \item \textbf{GKR4NLI} \citep{kalouli2020b} A system that uses Graphical Knowledge Representations (GKR). It converts sentences into semantic graphs capturing conceptual, contextual, lexical, and morphosyntactic constraints, aligns matching terms between sentence pairs, and computes the inference relation based on the specificity and instantiability of aligned terms.

    \item \textbf{Ccg2lambda1 and Ccg2lambda2} \citep{yanaka-etal-2018-acquisition}: Two variants of the ccg2lambda system, which maps inference pairs to logical formulas based on neo-Davidsonian event semantics. Both variants parse sentences into CCG (Combinatory Categorial Grammar) representations, convert them into semantic representations (directed acyclic graphs), and use a theorem prover with Natural Deduction rules. The difference lies in the CCG parser used: ccg2lambda1 employs the C\&C parser, while ccg2lambda2 uses the Easy-CCG parser.
    \item \textbf{LanGPro} \citep{abzianidze-2017-langpro}: A natural language theorem prover based on CCG and $\lambda$-terms, distinguished by its use of the analytic tableau method and natural logic rules for proofs. Its workflow includes a logical form generator (LLFgen), an optional text fragment aligner, and a first-order logic theorem prover with approximately 50 hand-coded rules.
\end{itemize}

\label{sssec:metricas}
We report standard metrics for classification tasks: Accuracy, Precision, Recall, and F1-score. For the SICK dataset (3 classes), we present the macro and weighted averages of these metrics, as well as the values per class. For SICK C-E (2 classes), we report the values per class and the overall accuracy.

\subsection{Results}
\label{subsec:resultados_cuantitativos}

Table~\ref{tab:resultados_principales} summarizes the performance of our proposed model on the SICK \cite{marelli-etal-2014-sick}.%

\begin{table}[htbp]
\centering
\resizebox{\textwidth}{!}{%

\begin{tabular}{lcccccc}
\toprule
\textbf{Corpus} & \textbf{Class} & \textbf{Precision} & \textbf{Recall} & \textbf{F1-score} & \textbf{Accuracy} & \textbf{Support} \\
\midrule
\multirow{3}{*}{SICK} & Contradiction & 0.85 & 0.78 & 0.81 & \multirow{3}{*}{0.83} & 712 \\
                       & Entailment    & 0.77 & 0.79 & 0.78 & & 1404 \\
                       & Neutral       & 0.85 & 0.86 & 0.85 & & 2790 \\
                       \cmidrule{2-7}
                       & Macro Avg     & 0.82 & 0.81 & 0.82 & & 4906 \\
\bottomrule
\end{tabular}
}

\caption{Performance of our model SLITE on the SICK test set. Numbers rounded to two decimal places.}
\label{tab:resultados_principales}
\end{table}

For a direct and fair comparison with the work of \citep{souza2025hybrid}, whose main reported metric is for the entailment class, we present in Table~\ref{tab:comparacion_sickce} the detailed results for that class on the SICK-CE, as well as the overall accuracy.

\begin{table}[htbp]
\centering
\resizebox{\textwidth}{!}{%
\begin{tabular}{lcccc}
\toprule
\textbf{Models} & \textbf{Accuracy} & \textbf{Precision (E)} & \textbf{Recall (E)} & \textbf{F1-score (E)} \\
\midrule
\textbf{SLITE} & \textbf{0.96} & \textbf{0.96} & \textbf{0.98} & \textbf{0.97} \\
\midrule
IsoLex  & 0.92 & 0.91 & 0.97 & 0.94 \\
RoBERTa (SotA) & 0.98 & 0.99 & 0.98 & 0.99 \\
\bottomrule
\end{tabular}
}

\caption{Performance comparison on the SICK C-E test suite, focusing on the \textbf{ENTAILMENT} class. The results for IsoLex and RoBERTa are taken from \citep{souza2025hybrid}.}
\label{tab:comparacion_sickce}
\end{table}

Table ~\ref{tab:comparacion_global_decimales} presents the results of a comparison between different semantic annotation frameworks.

\begin{table}[h!]
\centering
\resizebox{\textwidth}{!}{%
\begin{tabular}{lccc}
\toprule
\textbf{Models} & \textbf{SICK} & \textbf{\citep{hu-etal-2020-monalog}'s SICK} & \textbf{\citep{kalouli-etal-2023-curing} SICK} \\
\midrule
SLITE         & 0.828 & 0.823 & 0.824  \\
SLITE + Bert       & \textbf{0.833(+0.005)} & 0.818(-0.005) & 0.827(+0.003) \\
\midrule
Monalog          & 0.772 & 0.817 & —      \\
Monalog  + Bert        &  & \textbf{0.86(+0.043)} & —      \\
\midrule
Gkr4nli          & 0.785 & —     & 0.821  \\
Gkr4nli + Bert         &  & —     & \textbf{0.848(+0.027)}  \\
\bottomrule
\end{tabular}}\caption{Overall comparison of the accuracy of different systems on datasets derived from SICK. The references for the systems are: \cite{monalog} for Monalog and \cite{kalouli2020b} for Gkr4nli. Values originally published as percentages have been converted to decimal scale.}
\label{tab:comparacion_global_decimales}
\begin{minipage}{\textwidth}
\end{minipage}
\end{table}

The NLI symbolic systems—MonaLog or GKRNLI—return a neutral class when they cannot find sufficient evidence to derive a logically valid proof. Rather than accepting this neutral result as definitive, the system refers the decision to BERT, allowing the neural component to act as an oracle that resolves indecisive cases. This was implemented in the same way in our proposal for comparison purposes.

To evaluate our model's ability to capture more complex semantic phenomena, we compared it with various systems in the Logic and Commonsense datasets \citep{kalouli-etal-2023-curing}, which represent more challenging semantic annotations than the original SICK labels. Tables \ref{tab:bert_logica_sentido} and \ref{tab:comparacion_logica_sentido} presents the results.

\begin{table}[h]
\resizebox{\textwidth}{!}{%
\begin{tabular}{lclclcl}
   & \multicolumn{1}{l}{}         &                   & \multicolumn{1}{l}{}        &                 & \multicolumn{1}{l}{}           &                     \\ \hline
\textbf{Test set}                                        & \multicolumn{2}{c}{\textbf{Original}} & \multicolumn{2}{c}{\textbf{Logic}} & \multicolumn{2}{c}{\textbf{Commonsense}} \\
  & \multicolumn{1}{l}{Bert}     & SLITE            & \multicolumn{1}{l}{Bert}    & SLITE          & \multicolumn{1}{l}{Bert}       & SLITE              \\ \hline
Logic == Original        & \textbf{88.7}                         & 85.9              & \textbf{88.3}                        & 85.0            & \textbf{87.5}                           & 85.0                \\
Commonsense == Original & \textbf{88.6}                         & 85.9              & \textbf{88.1}                        & 84.8            & \textbf{87.5}                           & 84.9                \\
Logic != Original        & 20.1                         & \textbf{24.8}     & 29.4                        & \textbf{50.8}   & 26.1                           & \textbf{50.0}       \\
Commonsense != Original & 18.8                         & \textbf{22.7}     & 26.5                        & \textbf{49.6}   & 25.8                           & \textbf{49.0} \\
\bottomrule
\end{tabular}
}\caption{Accuracy of the pretrained models when fine-tuned and tested on different train and test sets.}
\label{tab:bert_logica_sentido}
\end{table}

\begin{table}[h!]
\centering
\begin{tabular}{lcc}
\toprule
\textbf{Models} & \textbf{Logic} & \textbf{Commonsense} \\
\midrule
SLITE        & \textbf{0.813} & \textbf{0.812} \\
\midrule
Gkr4nli          & 0.784 & 0.775 \\
\midrule
Monalog           & 0.743 & 0.725 \\
\midrule
Ccg2lambda2       & 0.763 & 0.744 \\
Ccg2lambda1        & 0.756 & 0.740 \\
LanGPro             & 0.755 & 0.740 \\
\bottomrule
\end{tabular}\caption{Overall comparison of accuracy of different systems in Logic and Commonsense derived from SICK. All values are expressed to three decimal places. The references for the systems are: \cite{monalog} for Monalog, \cite{kalouli2020b} for gkr4nli, and \cite{kalouli-etal-2023-curing} for the rest. }
\label{tab:comparacion_logica_sentido}
\begin{minipage}{\textwidth}
\end{minipage}
\end{table}

\subsection{Ablation Study}
\label{subsec:ablation}
\subsubsection{SICK}
To validate the contribution of each component of our feature set, we conducted a comprehensive ablation study. The objective is threefold: first, to quantify the marginal contribution of features by layer; second, to assess the relative importance of entity-level analyses versus lexical-level analyses; and third, to understand how features contribute to performance in specific classes.

We designed four ablation configurations based on disjoint groups of features, as defined in Table \ref{tab:features_summary}. All variants were evaluated on the complete SICK test set (three classes) using the identical experimental setup described in Section~\ref{subsec:configuracion}. Table~\ref{tab:ablation_complete} presents the comprehensive results of this analysis, including per-class F1-scores to enable fine-grained analysis of each configuration's strengths and weaknesses.

\begin{table}[h]
\centering
\small
\begin{tabular}{lccccccc}
\toprule
\textbf{Configuration} & \textbf{\textit{n\_features}} & \textbf{Accuracy} & \textbf{F1 (E)} & \textbf{F1 (C)} & \textbf{F1 (N)}\\
\midrule
\textbf{\textit{all (Full Model)}} & \textbf{17} & \textbf{0.828} & \textbf{0.783} & \textbf{0.812} & \textbf{0.854}\\
\midrule
\textit{w/o entity level} & 9 & 0.809 & 0.769 & 0.764 & 0.840\\
\textit{w/o DI layer} & 8 & 0.796 & 0.755 & 0.747 & 0.828\\
\textit{w/o lexical level} & 8 & 0.791 & 0.760 & 0.715 & 0.826\\
\textit{w/o SR layer} & 9 & 0.782  & 0.768 & 0.661 & 0.820\\
\bottomrule
\end{tabular}
\caption{Complete ablation study results on the SICK test set. F1 (E), F1 (C), and F1 (N) denote F1-scores for the entailment, contradiction, and neutral classes, respectively.}
\label{tab:ablation_complete}
\end{table}

\begin{figure}[h!]
    \centering
    \includegraphics[width=0.75\textwidth]{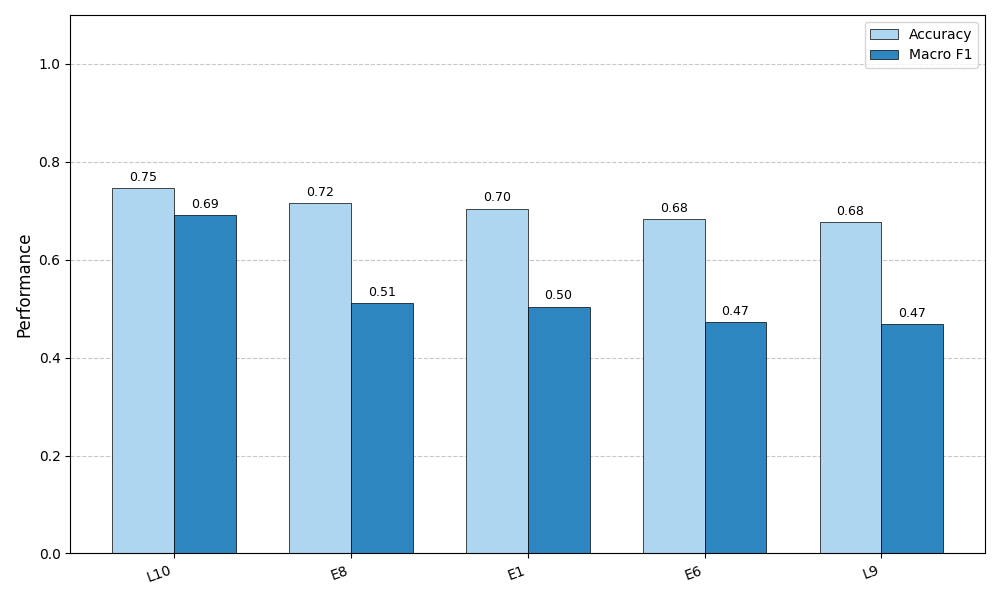}
    \caption{Performance analysis for the top five features on the  SICK corpus. The feature \textbf{L10} stands out for maintaining superior robustness in the Macro F1-score.}\label{fig:top5_sick}
\end{figure}

In Figure \ref{fig:top5_sick}, we contrast Accuracy with the Macro F1-score. In this three-class scenario (\textit{Entailment}, \textit{Neutral}, and \textit{Contradiction}), the Macro F1-score serves as a critical measure of stability. It evaluates how well each feature individually distinguishes between the three classes or whether it simply leans toward the most frequent categories. By presenting both metrics, we highlight the reliability of each feature in a highly complex inference environment.%

\subsubsection{SICK C-E}
\label{subsec:ablation_binary}

To complement the three-class ablation analysis, we conducted a focused ablation study on the SICK C-E binary classification task. Removing the neutral class allows us to isolate the model's capacity to distinguish between entailment and contradiction—the core distinction in RTE.

We evaluate the same four configurations defined previously, now on the 2,116-pair test set. Table~\ref{tab:ablation_binary} presents the complete results, including per-class F1-scores.

\begin{table}[htbp]
\centering
\begin{tabular}{lcccccc}
\toprule
\textbf{Configuration} & \textbf{No. features} & \textbf{Accuracy}  & \textbf{F1 (E)} & \textbf{F1 (C)}\\
\midrule
\textbf{\textit{all} (Full Model)} & \textbf{17} & \textbf{0.958}  & \textbf{0.969} & \textbf{0.937}\\
\midrule
\textit{w/o entity level} & 9 & 0.957 & 0.968 & 0.935\\
\textit{w/o DI layer} & 8 & 0.955 & 0.966 & 0.932\\
\midrule
\textit{w/o lexical level} & 8 & 0.927 &  0.945 & 0.892\\
\textit{w/o SR layer} & 9 & 0.921 & 0.941 & 0.880\\
\bottomrule
\end{tabular}
\caption{Ablation study results on the SICK C-E test set (binary classification). F1 (E) and F1 (C) denote F1-scores for entailment and contradiction classes.}
\label{tab:ablation_binary}
\end{table}

\begin{figure}[h!]
    \centering
    \includegraphics[width=0.75\textwidth]{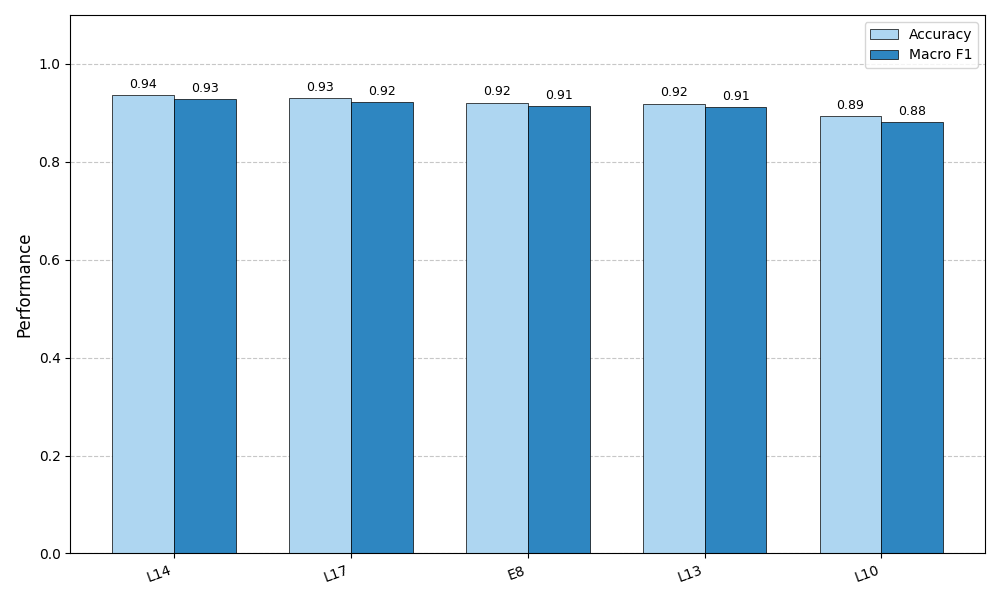}
    \caption{Performance analysis for the top five features on the  SICK-CE corpus. A high degree of consistency is observed between Accuracy and the Macro F1-score.}\label{fig:top5_sickce}
\end{figure}

Figure \ref{fig:top5_sickce} displays the performance of the five features that achieved the highest results individually within the SICK-CE corpus. The chart includes both Accuracy and the Macro F1-score to provide a comprehensive view of their predictive capabilities. This ensures that the performance of each individual feature reflects the feature’s ability to identify both \textit{Entailment} and \textit{Contradiction} with equal importance, regardless of potential class imbalances.%

\subsection{Qualitative Analysis}
\label{subsec:qualitative}

To complement the quantitative evaluation, we conduct a detailed qualitative analysis of individual predictions using SHAP (SHapley Additive exPlanations) \citep{lundberg2017unified}. SHAP provides a unified framework for interpreting model predictions by attributing each feature's contribution to the final decision.
This analysis serves three purposes: first, to validate that the model's decisions are grounded in linguistically meaningful patterns; second, to understand the specific failure modes that lead to misclassifications; and third, to illustrate how SHAP values can provide instance-level explanations for model behavior. Positive SHAP values push the prediction toward entailment; negative values push it toward contradiction.

We have selected four representative examples from SICK’s test suite that illustrate different phenomena: a correct prediction with a high level of confidence, an incorrect prediction with a high level of confidence, generalization in the presence of distractors, and a contradiction based on negation.\\

\begin{figure}[htbp]
     \centering
     \begin{subfigure}[b]{0.48\textwidth}
         \centering
         \includegraphics[width=\textwidth]{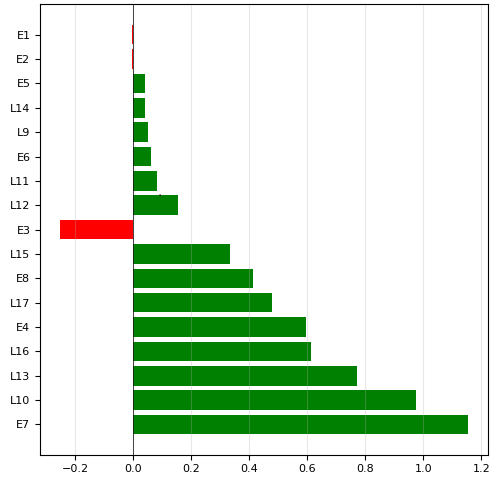}
         \caption{Case 1: Correct Prediction}
         \label{fig:shap_52}
     \end{subfigure}
     \hfill
     \begin{subfigure}[b]{0.48\textwidth}
         \centering
         \includegraphics[width=\textwidth]{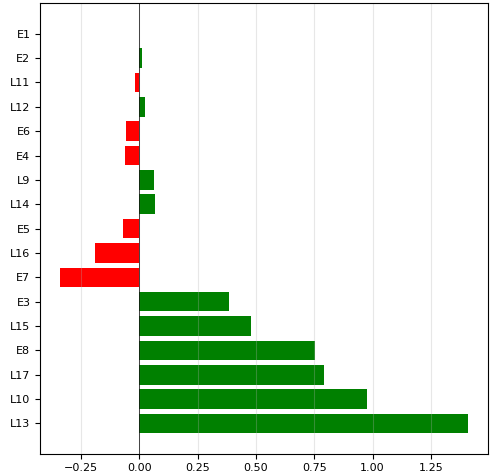}
         \caption{Case 2: Incorrect Prediction}
         \label{fig:shap_1435}
     \end{subfigure}

     \vspace{10pt}

     \begin{subfigure}[b]{0.48\textwidth}
         \centering
         \includegraphics[width=\textwidth]{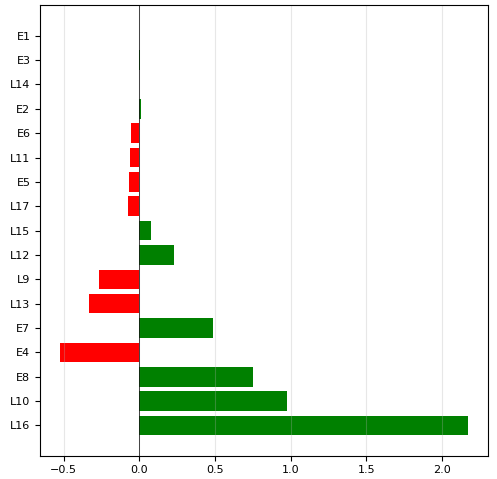}
         \caption{Case 3:  Generalization with Distractors}
         \label{fig:shap_395}
     \end{subfigure}
     \hfill
     \begin{subfigure}[b]{0.48\textwidth}
         \centering
         \includegraphics[width=\textwidth]{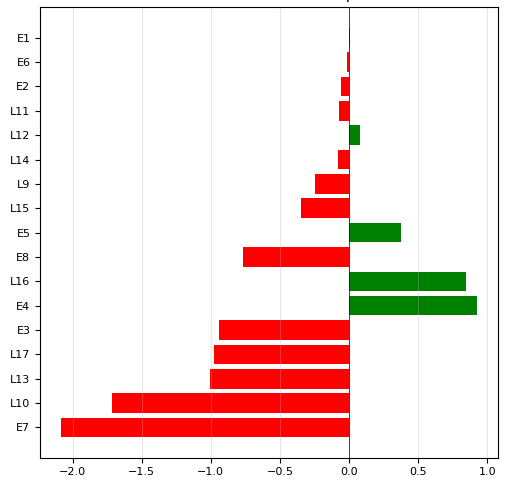}
         \caption{Case 4: The Negation-Driven Contradiction}
         \label{fig:shap_250}
     \end{subfigure}

     \caption{Qualitative analysis of feature contributions using SHAP additive explanations for four representative cases. Each plot illustrates how specific features push the model's output toward a specific  class. Green bars indicate features that increase the probability of the predicted entailment, while red bars represent counter-evidence.}
     \label{fig:grid_features}
\end{figure}

\begin{itemize}[nosep]
    \item[\textbf{Case 1:}] \textbf{High-Confidence Correct Prediction}
    \item[] \textbf{Premise:} A motorcyclist with a red helmet is riding a blue motorcycle down the road.
    \item[] \textbf{Hypothesis:} A motorcyclist is riding a motorbike along a roadway.
    \item[] \textbf{True label:} ENTAILMENT
    \item[] \textbf{Prediction:} ENTAILMENT (probability: 0.999)
\end{itemize}
\vspace{1em}

Figure~\ref{fig:shap_52} presents the waterfall plot for this instance, with base value $E[f(X)] = 1.117$ and final prediction $f(x) = 6.623$.
The SHAP analysis reveals that all major features contribute positively, with no conflicting signals.
The feature contributions drive the final prediction to $f(x) = 6.623$. The main features are: \textbf{E7} (+1.152) from entity-level captures entropy after removing specificity, while \textbf{L10} (+0.976) from lexical-level reflects low alignment of non-equivalent, non-specific words, signaling absence of contradictions. Among SR layer features, \textbf{L13} (+0.773) and \textbf{L16} (+0.612) capture directional semantic similarity and asymmetry, whereas entity-level features \textbf{E1–E2} contribute marginally. \textbf{E3} (-0.250)  is the only negative signal suggesting that high alignment without opposition may indicate redundancy rather than entailment evidence.\\
\vspace{1em}

\begin{itemize}[nosep]
    \item[\textbf{Case 2:}] \textbf{High-Confidence Incorrect Prediction}
    \item[] \textbf{Premise:} A woman in a red dress is putting away an instrument.
    \item[] \textbf{Hypothesis:} A woman in a red dress is playing an instrument.
    \item[] \textbf{True label:} CONTRADICTION
    \item[] \textbf{Prediction:} ENTAILMENT (probability: 0.995)
\end{itemize}
\vspace{1em}

Figure~\ref{fig:shap_1435} illustrates this striking error.
The feature contributions raise the final prediction to $f(x) = 5.3418$, yielding an erroneous Entailment prediction with probability 0.995. It is essential to note that no feature of the model provides sufficient evidence of a contradiction with respect to the others.

The features of the SR layer (\textbf{L13}: +1.407, \textbf{L17}: +0.791, \textbf{L15}: +0.480, \textbf{E8} (+0.754) and \textbf{L13}) drives the error, capturing semantic similarity between \textit{instrument} and \textit{playing}—actions that are lexically close (synonyms according to ConceptNet) due to stronger mental associations and a very high rate of semantic co-occurrence in the real world. However, there is alignment noise when using ConceptNet. The feature of the DI layer at the lexical-level contributes via \textbf{L10} (+0.976). Critically, the DI layer  provides negligible counter-signals (< 0.5), and only \textit{L16} feature captures the world-knowledge distinction between the two activities.\\
\newpage
\begin{itemize}[nosep]
    \item[\textbf{Case 3:}] \textbf{The Generalization with Distractors}
    \item[] \textbf{Premise:} The microphone in front of the talking parrot is being muted.
    \item[] \textbf{Hypothesis:} A parrot is speaking.
    \item[] \textbf{True label:} ENTAILMENT
    \item[] \textbf{Prediction:} ENTAILMENT
\end{itemize}

\vspace{1em}

The model's base value is $E[f(X)] = 1.117$, and feature contributions raise the final prediction to $f(x) = 4.421$, yielding a correct ENTAILMENT prediction with probability 0.988 (see Figure \ref{fig:shap_395}). The feature \textbf{L16} (+2.170) as the highest contributor—its strongly positive value (2.455) favoring entailment, capturing the directional asymmetry where the reverse similarity $H \to P$ is low because the hypothesis generalizes a specific element of the premise (\textit{talking parrot} $\to$ \textit{parrot is speaking}). The feature \textbf{L10} (+0.976), again showing its behavior with a positive value (0.96) favoring entailment. The features \textbf{E8} (+0.754) and \textbf{E7} (+0.483) favor entailment, while \textbf{E4} (-0.522) uniquely opposes it—suggesting that information loss when removing equivalent and opposition entities creates a counter-signal, though outweighed by other features. \\

\begin{itemize}[nosep]
    \item[\textbf{Case 4:}] \textbf{The Negation-Driven Contradiction}
    \item[] \textbf{Premise:} There is no man walking outside.
    \item[] \textbf{Hypothesis:} A man is walking outside.
    \item[] \textbf{True label:} CONTRADICTION
    \item[] \textbf{Prediction:} CONTRADICTION
\end{itemize}

\vspace{1em}

This example illustrates a clear contradiction signaled by explicit negation (see Figure~\ref{fig:shap_250}). The premise contains the negation marker ``no'', which directly contradicts the positive assertion in the hypothesis.

The feature contributions drive the final prediction downward to $f(x) = -4.9522$, yielding a correct Contradiction prediction with probability 0.993. All features contribute negatively, favoring contradiction. Although the specificity relation \textit{walk} $\to$ \textit{outside walk} was an artifact of imperfect entity construction, its contribution was outweighed by the strong opposition signal. The features of DI layer provides the strongest signal, with \textbf{E7} (-2.085) leading, its highly negative value (-3.823) indicating that entropy after removing specificity relations strongly signals contradiction. The feature \textbf{L10} (-1.713) also contributes to the contradiction demonstrating bimodal behavior where the same feature signals entailment or contradiction depending on context.
The SR features reinforces the contradiction with \textbf{L13} (-1.005), \textbf{E8} (-0.720) and \textbf{L17} (-0.975) capturing the directional mismatch between the negated premise and affirmative hypothesis, while \textbf{L14} and \textbf{L15} explicitly encodes negation. The features \textbf{E3} (-0.942), now favors contradiction—contrasting with the correct entailment case where E3 contributed negatively.

\subsection{Global Feature Contribution Analysis}
\label{subsec:shap_analysis}

While the quantitative results demonstrate the competitive performance of SLITE, the scientific value of a hybrid model resides in its inherent interpretability.

\begin{figure}[h]
\centering
\includegraphics[width=\textwidth]{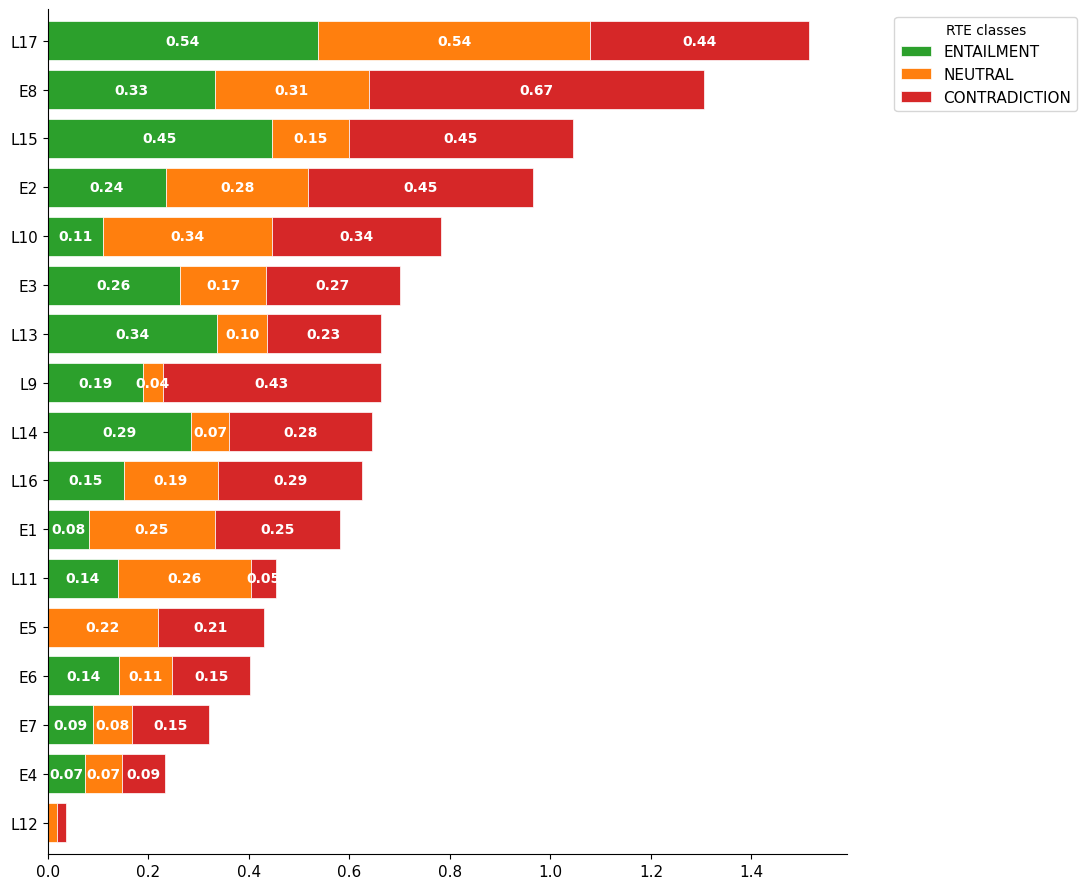}
\caption{\textbf{Global Feature Importance via Mean Absolute SHAP Values.}
    This figure illustrates the global distribution of SHAP values for the 17-feature set in the SLITE model on the SICK corpus.
    The horizontal axis represents the mean absolute SHAP value, quantifying the relative contribution of each feature to the classification of \textbf{Entailment} (green), \textbf{Neutral} (orange), and \textbf{Contradiction} (red).
    Features are ranked by their aggregate impact across all classes.}
\label{fig:shap_contributions}
\end{figure}

To deconstruct the model's decision process and
empirically validate the influence of our two
layers of analysis, we utilize SHAP summary plots
(Figure~\ref{fig:shap_contributions}). This global
analysis confirms that the model's predictive power
is grounded in the structured reasoning made
explicit by design, rather than in the exploitation
of superficial dataset artifacts. We present an
overview of global feature importance and the
complementary role that the Structural-Relational
and Distributional-Informational feature groups
play in distinguishing between the three inference
classes, providing the empirical foundation for
the qualitative analysis in
Subsection~\ref{subsec:discussion}.

\subsection{Discussion}
\label{subsec:discussion}

\paragraph{\textbf{Performance on RTE}}
A thorough evaluation of an RTE model requires
not only reporting performance metrics but also
examining the quality and biases of the evaluation
corpus. The SICK dataset, despite being a widely
used benchmark, presents characteristics that
restrict absolute interpretations of results.
SLITE is explicitly designed to capture a strong
directional flow of information from $P$ to $H$,
which makes it particularly sensitive to borderline
cases where the inference relation is ambiguous
or inconsistently annotated. This sensitivity
partly explains the performance gap between
SICK C-E (96\%) and the full SICK corpus (83\%),
since the neutral class is inherently heterogeneous,
encompassing both genuinely independent pairs and
those with plausible but uncertain entailment
relations, as discussed in Section~\ref{sec:notacion}.

Table~\ref{tab:resultados_principales} shows that
SLITE achieves 83\% overall accuracy and a weighted
F1-score of 0.83, with particularly strong
performance on the neutral class (F1 = 0.85),
consistent with the complementary role of the
Distributional-Informational layer in detecting
residual coherence. For the entailment class
(Table~\ref{tab:comparacion_sickce}), SLITE
attains an F1-score of 0.97, precision of 0.96,
and recall of 0.98, validating that the
Structural-Relational layer captures entailment
patterns accurately and in a balanced manner
across both classes.

\paragraph{\textbf{Ablation study}}
Table~\ref{tab:ablation_complete} shows that the full model achieves 0.828 accuracy with balanced per-class F1-scores. Removing SR layer causes the most severe degradation in contradiction detection (F1~(C) from 0.812 to 0.661, $-$15.1
points), while removing DI layer produces moderate but consistent degradation across all classes. Removing lexical-level features leads to the largest overall accuracy drop (to 0.791), confirming the foundational role of lexical signals.

The binary ablation on SICK C-E (Table~\ref{tab:ablation_binary}) reveals a clear asymmetry: removing SR layer causes a substantial drop (accuracy 0.958 to 0.921, contradiction F1 from 0.937 to 0.880), while removing DI layer produces negligible
degradation ($-$0.3 points). This confirms that in the absence of the neutral class, SR layer alone capture most of the available inferential signal, with DI layer contributing most distinctively to neutral class detection. This is confirmed by two
complementary findings. On the one hand, their removal produces a loss of 2.6 points in neutral F1 in three-class SICK (Table~\ref{tab:ablation_complete}). On the other hand, at the individual level, L10 ---the most stable DI feature (accuracy: 0.75, macro F1: 0.69,
Figure~\ref{fig:top5_sick})--- captures residual distributional coherence after removing semantically compatible terms, suggesting it is a primary contributor to this group-level effect.

The individual feature analysis (Figures~\ref{fig:top5_sick}
and~\ref{fig:top5_sickce}) reveals a sharp contrast between the two settings. In SICK C-E, SR lexical features dominate: L14 (accuracy: 0.94, macro F1: 0.93) and L17 (0.93, 0.92) lead the ranking, with a minimal accuracy-to-F1 gap across the top five features, indicating reliable discrimination between entailment and contradiction without class imbalance effects. In three-class SICK, however, most features exhibit a pronounced divergence between accuracy and macro F1. E8 (0.72 vs. 0.51) and E1
(0.70 vs. 0.50) maintain apparent accuracy above 0.70 but show substantially lower macro
F1, indicating bias toward the more frequent neutral and entailment classes and reduced
discriminative power for contradiction in the presence of the neutral class. The
exception is L10, which achieves the highest individual accuracy in three-class SICK (0.75) with the smallest accuracy-to-F1 gap (0.75
vs. 0.69), demonstrating that measuring residual alignment after removing semantically
compatible terms captures the inherent uncertainty of the neutral class without
sacrificing contradiction recognition.

The ablation results support three conclusions. First, SR layer constitute the essential core of SLITE, particularly for the entailment-contradiction axis. Second, DI layer act as specialized augmentations most needed in three-class settings, where the neutral class demands sensitivity to residual coherence that SR layer alone cannot fully provide. Third, the relative contribution of each dimension is task-dependent: SR layer dominate in binary classification, while the full integration of both dimensions is required for robust three-class performance. This complementarity validates our hypothesis that robust entailment recognition requires an ensemble of linguistically motivated signals operating across different dimensions of semantic analysis.

\paragraph{\textbf{Comparison with other models}}
Table~\ref{tab:comparacion_sickce} shows that SLITE
(accuracy: 0.96, F1: 0.97) outperforms IsoLex
(0.92, 0.94) by four percentage points in accuracy
and three in F1, with precision (0.96) and recall
(0.98) matching or exceeding those of IsoLex (0.91
and 0.97, respectively). As expected, it does not
surpass fine-tuned RoBERTa (accuracy: 0.98, F1:
0.99); however, the gap is remarkably small given
the extreme asymmetry in model complexity: SLITE
uses a logistic regression classifier on 17 interpretable features derived from static embeddings, whereas RoBERTa is an end-to-end transformer with hundreds of millions of parameters.

Table~\ref{tab:comparacion_global_decimales} shows
that SLITE achieves 0.828 accuracy on SICK,
substantially above Monalog (0.772) and Gkr4nli
(0.785). On the Monalog corpus \citep{monalog},
SLITE scores 0.823, slightly above base Monalog
(0.817); on Kalouli \citep{kalouli2020b}, it
reaches 0.824, close to Gkr4nli (0.821).
Crucially, incorporating BERT does not
fundamentally alter SLITE's behavior ($+$0.005),
in sharp contrast with Monalog ($+$0.043) and
Gkr4nli ($+$0.027). This stability across datasets suggests that the Structural-Relational and Distributional-Informational features capture fundamental aspects of textual entailment that transcend the particularities of individual semantic re-annotations, rather than exploiting corpus-specific distributional patterns.

Table~\ref{tab:bert_logica_sentido} presents
cross-dataset results under annotation shift,
where training and test sets follow different
labeling criteria. When trained on logic-annotated data and tested on original labels ---the condition that most directly exposes reliance on corpus-specific biases--- SLITE achieves 50.8\% accuracy against BERT's 29.4\%, a gap of 21.4 percentage points. The reverse condition (trained on original, tested on logic-mismatched) yields 24.8\% for SLITE versus 20.1\% for BERT, a smaller but consistent advantage. An analogous pattern holds for the commonsense condition (49.0\% vs. 25.8\% and 22.7\% vs. 18.8\%). These results indicate that BERT overfits to
the labeling biases of its training data and
fails to generalize when the inference criterion
changes, while SLITE's linguistically motivated
features provide more robust generalization
across annotation frameworks.

Table~\ref{tab:comparacion_logica_sentido} shows
that SLITE achieves 0.813 on the Logic corpus
and 0.812 on the Commonsense corpus, outperforming all reference systems in both cases. The advantage over Gkr4nli reaches $+$2.9 points on Logic and $+$3.7 on Commonsense; over Monalog, $+$7.0 and $+$8.7 points respectively. Systems such as ccg2lambda and LanGPro show significant drops on these more demanding corpora, whereas SLITE maintains stable and consistently superior performance. As noted in Section~\ref{sec:trabajo_relacionado},
the Logic and Commonsense corpora were designed
to evaluate precisely the inferential phenomena
that Natural Logic-based systems are expected to
handle well. The fact that SLITE outperforms
Monalog and Gkr4nli on these corpora by
substantial margins suggests that dependency-based entity analysis combined with Distributional-Informational features captures inferential structure that monotonicity-based approaches systematically miss; a consequence of operating at a finer level of semantic granularity than the monotonicity calculus affords.

\paragraph{\textbf{Qualitative analysis}}
Case~1 illustrates a classic entailment where the hypothesis generalizes the premise: \textit{motorcyclist} $\rightarrow$ \textit{motorcyclist}, \textit{blue
motorcycle} $\rightarrow$ \textit{motorbike} (hypernymy), and \textit{along a roadway} $\rightarrow$ \textit{down the road}. The model correctly identifies that specific details in the premise (red helmet, blue motorcycle) do
not affect the entailment relation, producing a
high-confidence correct prediction with all features contributing positively.

Case~2 reveals the primary limitation of the approach. The premise and hypothesis differ only in the verb phrase, \textit{putting away} versus
\textit{playing}, activities that are mutually
exclusive in the world but indistinguishable
through lexical-semantic relations alone.
Since no feature captures this world-knowledge
distinction, all features erroneously favor
entailment with high confidence, representing
the most challenging class of errors for our
framework: cases where world knowledge rather
than lexical-semantic structure determines
the inference relation.

Case~3 demonstrates the model's ability to handle structural complexity: it correctly identifies that \textit{talking parrot} entails \textit{parrot is speaking} despite the premise containing additional elements (microphone, muted) and a more complex syntactic structure.

Case~4 shows the complementary strength: when explicit negation is present (\textit{no man walking outside} vs. \textit{a man is walking outside}), all features converge on contradiction, demonstrating that negation signals propagate effectively through both Structural-Relational and Distributional-Informational features when captured
by the external knowledge resource.

The four cases reveal a consistent pattern: the
feature set operates through complementarity, as
no single feature or group is sufficient for robust entailment recognition, but their orchestrated combination produces coherent and diagnosable decisions. Features such as L10 and E3 exhibit bimodal behavior, favoring entailment in some contexts and contradiction in others, with their contribution direction determined by semantic context and feature value. Structural-Relational features capture lexical-semantic alignment and directional mismatch, while Distributional-Informational features capture residual coherence and entropy patterns; their interaction ensures that when one dimension provides ambiguous signals, the other provides corrective evidence. The exception, as Case~2 illustrates, occurs when the external knowledge resource lacks the relations required to distinguish between lexically similar
but logically incompatible predicates, a limitation that defines the boundary of the current framework and motivates the future directions outlined in Section~\ref{sec:conclusion}.

\paragraph{\textbf{Global contribution of features}} Figure~\ref{fig:shap_contributions} shows that L17 emerges as the most influential and balanced predictor (Entailment: 0.54, Neutral: 0.54, Contradiction: 0.44), confirming that the coverage of $P$ over $H$ is the primary indicator of inferential coherence. The high influence of L17 on the Neutral class suggests that the model uses the coverage gap as a reliable indicator that the hypothesis introduces unsupported information.

E8 serves as the primary switch for contradiction detection (SHAP: 0.67). Since E8 returns zero when it detects an opposition relation ($\mathcal{G}_{op}$) or negation, its high weight indicates that a single
localized contradiction in the semantic-structural graph is sufficient to override lexical similarity and trigger the Contradiction class.

A critical finding concerns the contrast between L9 and its knowledge-aware counterparts. L9 computes a global alignment by summing maximum cosine similarities between word embeddings in $P$ and $H$, without distinguishing between types of semantic overlap. Features that operate over semantic submatrices ---E5,
L10, and L11--- provide substantially higher predictive power for the Neutral and Contradiction classes precisely because they isolate the contribution of specific relation types rather than aggregating over
all similarity values indiscriminately. This contrast constitutes empirical support for the core design decision of the Distributional-Informational dimension: structured sub-representations of the similarity matrix are more informative than raw alignment.

Distributional-Informational features L12 and E4
show moderate global contributions. While L17 and E8 define the broad decision boundaries, L12 and E4 act as calibration mechanisms, quantifying directional information flow and contributing to the precision observed on SICK-CE.

The Jaro-based feature block (L13--L16) collectively provides a nuanced inferential signal that complements the coverage measure of L17. L15 (SHAP: 0.45 for Entailment) identifies cases where the hypothesis refines the premise's meaning through specificity matches ($sm=1$) without introducing contradictory information, demonstrating that the integration of hierarchical relations directly into the matching algorithm provides a more stable foundation for entailment detection than vector proximity alone. L14 (SHAP: 0.28 for Contradiction) acts as a precision filter for logical conflicts, detecting contradictory matches ($cm=1$) at the immediate semantic level ---explicit negation and antonyms that global embedding averages tend to conceal. L16 (Reverse Jaro, SHAP: 0.19 for Neutral) exploits the non-symmetry of entailment: evaluating the mapping $H \rightarrow P$ identifies cases where
the hypothesis introduces asymmetric information
that prevents a valid inference even under high
lexical overlap. Together, the contrast between
the predictive power of these Structural-Relational features and the negligible contribution of L9 supports a central claim of this work: for RTE, the quality of semantic alignment is far more consequential than the quantity of statistical similarity.

\section{Conclusions}
\label{sec:conclusion}
This paper presented a hybrid approach to Recognizing Textual Entailment motivated by a central methodological question: how can the semantic relations between a premise and a hypothesis be analyzed simultaneously from a structural-relational perspective ---grounded in semantic compatibility and incompatibility between compositional entities--- and from a distributional-informational perspective ---grounded in the patterns of information change between their embedding-based representations--- so as to produce an inferential account that is both competitive and transparent by design? Our answer integrates these two dimensions into a unified, non-serialized framework that avoids the error propagation of serialized pipelines while preserving the linguistic interpretability that opaque neural architectures sacrifice.

Our answer is the SLITE feature set: 17 features
organized across two layers of semantic analysis
and two levels of linguistic representation.
The Structural-Relational layer contributes two
groups of features: at the entity level, E1 and
E2 quantify the proportional weight of
compatibility and ambiguity relations between
the compositional entities of $P$ and $H$, while
E8 synthesizes these proportions into a weighted
entailment score that returns 1 when only
generality relations exist and 0 when any
opposition is detected; at the lexical level,
L13--L17 implement a semantically extended Jaro
measure with polarity-sensitive matching and
directional penalties. The Distributional-
Informational layer also contributes two groups:
at the entity level, E3--E7 measure residual
similarity and entropy over the submatrices
resulting from removing specific semantic groups;
at the lexical level, L9--L12 quantify residual
coherence and directional uncertainty reduction
over lexical submatrices, including transfer
entropy as a novel operationalization of
entailment as directional uncertainty reduction
from $P$ to $H$.

A logistic regression trained on these features achieves 96\% accuracy on SICK C-E and 83\% on three-class SICK. On SICK C-E, our model surpasses IsoLex by 4 points and comes within 2 points of RoBERTa; a remarkable result given that our model uses a linear classifier on 17 interpretable features versus RoBERTa's 355 million parameters.

SHAP-based analysis and ablation studies validate that the feature set operates through complementarity. The primary drivers of classification are the Structural-Relational features, particularly L17 (the single most predictive feature) and E8 (the second most predictive), which dominate in both three-class and binary settings. Distributional-Informational features contribute most distinctively to neutral class detection: their removal produces a loss of 2.6 points in neutral F1 in three-class SICK, and L10, the most stable individual DI feature,
captures residual distributional coherence after removing semantically compatible terms, suggesting it is a primary contributor to this group-level effect. In binary classification, SR features alone capture most of the available inferential signal, with DI features contributing negligibly ($-$0.3 points when removed), confirming that their distinctive role is tied to the inferential complexity introduced by the neutral class.

The analysis yields four main conclusions: (i) entailment is most reliably detected through Structural-Relational alignment signals, particularly the weighted interaction of semantic groups and lexical-semantic coverage; (ii) contradiction requires the full integration of lexical opposition, structural incompatibility, and distributional-informational directionality, as confirmed by the 15.1-point drop in contradiction F1 when SR features are removed; (iii) the neutral class justifies the full feature
set's complexity, as it is precisely where the interaction of both dimensions is most needed; and (iv) world knowledge gaps remain the primary limitation, defining the boundary condition where no feature combination can compensate for the absence of event-level mutual exclusivity relations.

We identify four priorities for future research. First, the model's primary failure mode ---world knowledge gaps, as illustrated by Case~2--- motivates the integration of commonsense knowledge beyond ConceptNet's current coverage, particularly for event-level mutual exclusivity relations not well represented in its graph. Second, implicit negation
expressed through adverbial or aspectual modifiers (\textit{almost}, \textit{barely}, \textit{stop}) represents a systematic source of error that calls for dedicated feature engineering targeting pragmatic phenomena. Third, the current entity alignment is
flat; a recursive extension that handles nested modification structures would extend coverage to more complex syntactic configurations. Finally, evaluation on adversarially constructed benchmarks such as ANLI \citep{nie-etal-2020-adversarial} and HANS \citep{Mccoy2019} remains a priority to assess robustness beyond the SICK family of datasets.

Our work demonstrates that thorough linguistic analysis and feature engineering remain viable for NLI and that the goals of accuracy and transparency need not be in conflict. By decomposing the semantic relations between $P$ and $H$ into interpretable sub-representations and operationalizing their information change as inferential evidence, our model achieves near-state-of-the-art performance with full explainability, challenging the assumption that massive transformers are needed to obtain competitive results. In domains where transparency is essential, characterizing what a model learns constitutes a scientific contribution in itself. We hope this work will encourage further exploration of hybrid approaches that combine symbolic precision with statistical learning, fostering a more productive dialogue between linguistic theory and computational modeling of inference.

\bibliographystyle{compling}
\bibliography{biblio}

\begin{thebibliography}{48}
\expandafter\ifx\csname natexlab\endcsname\relax\def\natexlab#1{#1}\fi

\bibitem[{Abzianidze(2017)}]{abzianidze-2017-langpro}
Abzianidze, Lasha. 2017.
\newblock {L}ang{P}ro: Natural language theorem prover.
\newblock In \emph{Proceedings of the 2017 Conference on Empirical Methods in
  Natural Language Processing: System Demonstrations}, pages 115--120,
  Association for Computational Linguistics, Copenhagen, Denmark.

\bibitem[{Amigó et~al.(2022)Amigó, Ariza-Casabona, Fresno, and
  Martí}]{amigo2022}
Amigó, Enrique, Alejandro Ariza-Casabona, Victor Fresno, and M.~Antònia
  Martí. 2022.
\newblock Information theory–based compositional distributional semantics.
\newblock \emph{Computational Linguistics}, 48(4):907--948.

\bibitem[{Banko et~al.(2007)Banko, Cafarella, Soderland, Broadhead, and
  Etzioni}]{Banko2007}
Banko, Michele, Michael~J. Cafarella, Stephen Soderland, Matt Broadhead, and
  Oren Etzioni. 2007.
\newblock Open information extraction from the web.
\newblock In \emph{Proceedings of the 20th International Joint Conference on
  Artificial Intelligence ({IJCAI})}, pages 2670--2676, Hyderabad, India.

\bibitem[{Belinkov(2022)}]{belinkov2022probing}
Belinkov, Yonatan. 2022.
\newblock Probing classifiers: Promises, shortcomings, and advances.
\newblock \emph{Computational Linguistics}, 48(1):207--219.

\bibitem[{Bhalla, Srinivas, and Lakkaraju(2023)}]{NEURIPS2023_89beb2a3}
Bhalla, Usha, Suraj Srinivas, and Himabindu Lakkaraju. 2023.
\newblock Discriminative feature attributions: Bridging post hoc explainability
  and inherent interpretability.
\newblock In \emph{Advances in Neural Information Processing Systems},
  volume~36, pages 44105--44122, Curran Associates, Inc.

\bibitem[{Bowman et~al.(2015)Bowman, Angeli, Potts, and
  Manning}]{bowman2015large}
Bowman, Samuel~R., Gabor Angeli, Christopher Potts, and Christopher~D. Manning.
  2015.
\newblock A large annotated corpus for learning natural language inference.
\newblock In \emph{Proceedings of the 2015 Conference on Empirical Methods in
  Natural Language Processing}, pages 632--642, Association for Computational
  Linguistics, Lisbon, Portugal.

\bibitem[{Burkart and Huber(2021)}]{bukart2021}
Burkart, Nadia and Marco~F. Huber. 2021.
\newblock A survey on the explainability of supervised machine learning.
\newblock \emph{J. Artif. Int. Res.}, 70:245–317.

\bibitem[{Cruse(2004)}]{cruse2004meaning}
Cruse, D.~Alan. 2004.
\newblock \emph{Meaning in Language: An Introduction to Semantics and
  Pragmatics}.
\newblock Oxford University Press, Oxford.

\bibitem[{Dagan, Glickman, and Magnini(2006)}]{dagan2006}
Dagan, Ido, Oren Glickman, and Bernardo Magnini. 2006.
\newblock The pascal recognising textual entailment challenge.
\newblock \emph{In Quionero-Candela et al., editor, LNAI 3944: MLCW2005}.

\bibitem[{Devlin et~al.(2019)Devlin, Chang, Lee, and
  Toutanova}]{devlin2019bert}
Devlin, Jacob, Ming-Wei Chang, Kenton Lee, and Kristina Toutanova. 2019.
\newblock {BERT}: Pre-training of deep bidirectional transformers for language
  understanding.
\newblock In \emph{Proceedings of the 2019 Conference of the North {A}merican
  Chapter of the Association for Computational Linguistics: Human Language
  Technologies, Volume 1 (Long and Short Papers)}, pages 4171--4186,
  Association for Computational Linguistics, Minneapolis, Minnesota.

\bibitem[{Engler et~al.(2022)Engler, Sikdar, Lutz, and
  Strohmaier}]{engler-etal-2022-sensepolar}
Engler, Jan, Sandipan Sikdar, Marlene Lutz, and Markus Strohmaier. 2022.
\newblock {S}ense{POLAR}: Word sense aware interpretability for pre-trained
  contextual word embeddings.
\newblock In \emph{Findings of the Association for Computational Linguistics:
  EMNLP 2022}, pages 4607--4619, Association for Computational Linguistics, Abu
  Dhabi, United Arab Emirates.

\bibitem[{Fellbaum(1998)}]{Fellbaum1998}
Fellbaum, Christiane, editor. 1998.
\newblock \emph{{WordNet}: An Electronic Lexical Database}.
\newblock Language, Speech, and Communication. The MIT Press, Cambridge, MA.

\bibitem[{Ferrando et~al.(2024)Ferrando, Sarti, Bisazza, and
  Costa-jussà}]{ferrando2024primer}
Ferrando, Javier, Gabriele Sarti, Arianna Bisazza, and Marta~R. Costa-jussà.
  2024.
\newblock A primer on the inner workings of transformer-based language models.

\bibitem[{Gururangan et~al.(2018)Gururangan, Swayamdipta, Levy, Schwartz,
  Bowman, and Smith}]{gururangan-etal-2018-annotation}
Gururangan, Suchin, Swabha Swayamdipta, Omer Levy, Roy Schwartz, Samuel Bowman,
  and Noah~A. Smith. 2018.
\newblock Annotation artifacts in natural language inference data.
\newblock In \emph{Proceedings of the 2018 Conference of the North {A}merican
  Chapter of the Association for Computational Linguistics: Human Language
  Technologies, Volume 2 (Short Papers)}, pages 107--112, Association for
  Computational Linguistics, New Orleans, Louisiana.

\bibitem[{Hu et~al.(2020{\natexlab{a}})Hu, Chen, Richardson, Mukherjee, Moss,
  and Kuebler}]{hu-etal-2020-monalog}
Hu, Hai, Qi~Chen, Kyle Richardson, Atreyee Mukherjee, Lawrence~S. Moss, and
  Sandra Kuebler. 2020{\natexlab{a}}.
\newblock {M}ona{L}og: a lightweight system for natural language inference
  based on monotonicity.
\newblock In \emph{Proceedings of the Society for Computation in Linguistics
  2020}, pages 334--344, Association for Computational Linguistics, New York,
  New York.

\bibitem[{Hu et~al.(2020{\natexlab{b}})Hu, Chen, Richardson, Mukherjee, Moss,
  and Kuebler}]{monalog}
Hu, Hai, Qi~Chen, Kyle Richardson, Atreyee Mukherjee, Lawrence~S Moss, and
  Sandra Kuebler. 2020{\natexlab{b}}.
\newblock {MonaLog: a Lightweight System for Natural Language Inference Based
  on Monotonicity}.
\newblock In \emph{Proceedings of SCiL}, pages 334--344.

\bibitem[{Hurford, Heasley, and Smith(2007)}]{hurford2007semantics}
Hurford, James~R., Brendan Heasley, and Michael~B. Smith. 2007.
\newblock \emph{Semantics: A Coursebook}.
\newblock Cambridge University Press, Cambridge.

\bibitem[{Iwamoto, Kohita, and Wachi(2021)}]{iwamoto-etal-2021-polar}
Iwamoto, Ran, Ryosuke Kohita, and Akifumi Wachi. 2021.
\newblock Polar embedding.
\newblock In \emph{Proceedings of the 25th Conference on Computational Natural
  Language Learning}, pages 470--480, Association for Computational
  Linguistics, Online.

\bibitem[{Jaro(1989)}]{Jaro1989}
Jaro, Matthew~A. 1989.
\newblock Advances in record-linkage methodology as applied to matching the
  1985 census of tampa, florida.
\newblock \emph{Journal of the American Statistical Association},
  84(406):414--420.

\bibitem[{Jeffries(1998)}]{jeffries1998meaning}
Jeffries, Lesley. 1998.
\newblock \emph{Meaning in English: An Introduction to Language Study}.
\newblock Bloomsbury Publishing, London.

\bibitem[{Kaiser and Schreiber(2002)}]{KAISER200243}
Kaiser, A. and T.~Schreiber. 2002.
\newblock Information transfer in continuous processes.
\newblock \emph{Physica D: Nonlinear Phenomena}, 166(1):43--62.

\bibitem[{Kalouli, Crouch, and de~Paiva(2020)}]{kalouli2020b}
Kalouli, Aikaterini-Lida, Richard Crouch, and Valeria de~Paiva. 2020.
\newblock Hy-nli: a hybrid system for natural language inference.
\newblock In \emph{Proceedings of the 28th International Conference on
  Computational Linguistics}, COLING '20, page 5235–5249, Association for
  Computational Linguistics.

\bibitem[{Kalouli et~al.(2023)Kalouli, Hu, Webb, Moss, and
  de~Paiva}]{kalouli-etal-2023-curing}
Kalouli, Aikaterini-Lida, Hai Hu, Alexander~F. Webb, Lawrence~S. Moss, and
  Valeria de~Paiva. 2023.
\newblock Curing the {SICK} and other {NLI} maladies.
\newblock \emph{Computational Linguistics}, 49(1):199--243.

\bibitem[{Khot, Sabharwal, and Clark(2018)}]{Khot2018SciTaiLAT}
Khot, Tushar, Ashish Sabharwal, and Peter Clark. 2018.
\newblock Scitail: A textual entailment dataset from science question
  answering.
\newblock In \emph{AAAI Conference on Artificial Intelligence}.

\bibitem[{Liu et~al.(2019)Liu, Ott, Goyal, Du, Joshi, Chen, Levy, Lewis,
  Zettlemoyer, and Stoyanov}]{liu2019roberta}
Liu, Yinhan, Myle Ott, Naman Goyal, Jingfei Du, Mandar Joshi, Danqi Chen, Omer
  Levy, Mike Lewis, Luke Zettlemoyer, and Veselin Stoyanov. 2019.
\newblock Roberta: {A} robustly optimized {BERT} pretraining approach.
\newblock \emph{CoRR}, abs/1907.11692.

\bibitem[{Lundberg and Lee(2017)}]{lundberg2017unified}
Lundberg, Scott~M and Su-In Lee. 2017.
\newblock A unified approach to interpreting model predictions.
\newblock In \emph{Advances in Neural Information Processing Systems 30}. pages
  4765--4774.

\bibitem[{Lyons(1977)}]{lyons1977semantics}
Lyons, John. 1977.
\newblock \emph{Semantics}.
\newblock Cambridge University Press, Cambridge.

\bibitem[{MacCartney and Manning(2009{\natexlab{a}})}]{maccartney2009}
MacCartney, Bill and Christopher~D. Manning. 2009{\natexlab{a}}.
\newblock An extended model of natural logic.
\newblock In \emph{Proceedings of the Eight International Conference on
  Computational Semantics}, pages 140--156, Association for Computational
  Linguistics, Tilburg, The Netherlands.

\bibitem[{MacCartney and
  Manning(2009{\natexlab{b}})}]{maccartney-manning-2009-extended}
MacCartney, Bill and Christopher~D. Manning. 2009{\natexlab{b}}.
\newblock An extended model of natural logic.
\newblock In \emph{Proceedings of the Eight International Conference on
  Computational Semantics}, pages 140--156, Association for Computational
  Linguistics, Tilburg, The Netherlands.

\bibitem[{Marelli et~al.(2014)Marelli, Menini, Baroni, Bentivogli, Bernardi,
  and Zamparelli}]{marelli-etal-2014-sick}
Marelli, Marco, Stefano Menini, Marco Baroni, Luisa Bentivogli, Raffaella
  Bernardi, and Roberto Zamparelli. 2014.
\newblock A {SICK} cure for the evaluation of compositional distributional
  semantic models.
\newblock In \emph{Proceedings of the Ninth International Conference on
  Language Resources and Evaluation ({LREC}'14)}, pages 216--223, European
  Language Resources Association (ELRA), Reykjavik, Iceland.

\bibitem[{McCoy, Pavlick, and Linzen(2019)}]{Mccoy2019}
McCoy, R.~Thomas, Ellie Pavlick, and Tal Linzen. 2019.
\newblock Right for the wrong reasons: Diagnosing syntactic heuristics in
  natural language inference.
\newblock \emph{CoRR}, abs/1902.01007.

\bibitem[{Murphy(2003)}]{murphy2003semantic}
Murphy, M.~Lynne. 2003.
\newblock \emph{Semantic Relations and the Lexicon: Antonymy, Synonymy, and
  Other Paradigms}.
\newblock Cambridge University Press, Cambridge.

\bibitem[{Nie et~al.(2020)Nie, Williams, Dinan, Bansal, Weston, and
  Kiela}]{nie-etal-2020-adversarial}
Nie, Yixin, Adina Williams, Emily Dinan, Mohit Bansal, Jason Weston, and Douwe
  Kiela. 2020.
\newblock Adversarial {NLI}: A new benchmark for natural language
  understanding.
\newblock In \emph{Proceedings of the 58th Annual Meeting of the Association
  for Computational Linguistics}, Association for Computational Linguistics.

\bibitem[{Parikh et~al.(2016)Parikh, T{\"a}ckstr{\"o}m, Das, and
  Uszkoreit}]{parikh-etal-2016-decomposable}
Parikh, Ankur, Oscar T{\"a}ckstr{\"o}m, Dipanjan Das, and Jakob Uszkoreit.
  2016.
\newblock A decomposable attention model for natural language inference.
\newblock In \emph{Proceedings of the 2016 Conference on Empirical Methods in
  Natural Language Processing}, pages 2249--2255, Association for Computational
  Linguistics, Austin, Texas.

\bibitem[{Rocktäschel et~al.(2016)Rocktäschel, Grefenstette, Hermann,
  Kočiský, and Blunsom}]{rocktaschel2016reasoning}
Rocktäschel, Tim, Edward Grefenstette, Karl~Moritz Hermann, Tomáš Kočiský,
  and Phil Blunsom. 2016.
\newblock Reasoning about entailment with neural attention.

\bibitem[{Rudin(2019)}]{rudin2019}
Rudin, Cynthia. 2019.
\newblock Stop explaining black box machine learning models for high stakes
  decisions and use interpretable models instead.
\newblock \emph{Nature Machine Intelligence}, 1:206--215.

\bibitem[{Santus et~al.(2014)Santus, Lenci, Lu, and Schulte~im
  Walde}]{santus2014}
Santus, Enrico, Alessandro Lenci, Qin Lu, and Sabine Schulte~im Walde. 2014.
\newblock Chasing hypernyms in vector spaces with entropy.
\newblock In \emph{Proceedings of the 14th Conference of the {E}uropean Chapter
  of the Association for Computational Linguistics, volume 2: Short Papers},
  pages 38--42, Association for Computational Linguistics, Gothenburg, Sweden.

\bibitem[{Schreiber(2000)}]{PhysRevLett.85.461}
Schreiber, Thomas. 2000.
\newblock Measuring information transfer.
\newblock \emph{Phys. Rev. Lett.}, 85:461--464.

\bibitem[{Silva, Freitas, and Handschuh(2020)}]{vivianXTE}
Silva, Vivian Dos~Santos, Andr{\'{e}} Freitas, and Siegfried Handschuh. 2020.
\newblock {XTE:} explainable text entailment.
\newblock \emph{CoRR}, abs/2009.12431.

\bibitem[{Souza and Lopes(2025)}]{souza2025hybrid}
Souza, Rodrigo and Marcos Lopes. 2025.
\newblock A hybrid approach to natural language inference for the sick dataset.
\newblock \emph{Computer Speech and Language}, 90:101736.

\bibitem[{Speer, Chin, and
  Havasi(2018)}]{speer2018conceptnet55openmultilingual}
Speer, Robyn, Joshua Chin, and Catherine Havasi. 2018.
\newblock Conceptnet 5.5: An open multilingual graph of general knowledge.

\bibitem[{Stali{\={u}}nait{\.{e}} and
  Vlachos(2025)}]{staliunaite-vlachos-2025-uncertain}
Stali{\={u}}nait{\.{e}}, Ieva~Raminta and Andreas Vlachos. 2025.
\newblock Uncertain (mis)takes at {L}e{W}i{D}i-2025: Modeling human label
  variation with semantic entropy.
\newblock In \emph{Proceedings of the The 4th Workshop on Perspectivist
  Approaches to NLP}, pages 256--262, Association for Computational
  Linguistics, Suzhou, China.

\bibitem[{Torres-Moreno and Hermosillo-Valadez(2026)}]{TORRESMORENO2026114825}
Torres-Moreno, David and Jorge Hermosillo-Valadez. 2026.
\newblock Semantic knowledge abstraction: Consistent reasoning in large
  language models for natural language inference.
\newblock \emph{Knowledge-Based Systems}, 332:114825.

\bibitem[{Wang et~al.(2019)Wang, Pruksachatkun, Nangia, Singh, Michael, Hill,
  Levy, and Bowman}]{Wang2019SuperGLUEAS}
Wang, Alex, Yada Pruksachatkun, Nikita Nangia, Amanpreet Singh, Julian Michael,
  Felix Hill, Omer Levy, and Samuel~R. Bowman. 2019.
\newblock Superglue: A stickier benchmark for general-purpose language
  understanding systems.
\newblock \emph{ArXiv}, abs/1905.00537.

\bibitem[{Williams, Nangia, and Bowman(2018)}]{williams-etal-2018-broad}
Williams, Adina, Nikita Nangia, and Samuel Bowman. 2018.
\newblock A broad-coverage challenge corpus for sentence understanding through
  inference.
\newblock In \emph{Proceedings of the 2018 Conference of the North {A}merican
  Chapter of the Association for Computational Linguistics: Human Language
  Technologies, Volume 1 (Long Papers)}, pages 1112--1122, Association for
  Computational Linguistics, New Orleans, Louisiana.

\bibitem[{Yanaka et~al.(2018)Yanaka, Mineshima, Mart{\'i}nez-G{\'o}mez, and
  Bekki}]{yanaka-etal-2018-acquisition}
Yanaka, Hitomi, Koji Mineshima, Pascual Mart{\'i}nez-G{\'o}mez, and Daisuke
  Bekki. 2018.
\newblock Acquisition of phrase correspondences using natural deduction proofs.
\newblock In \emph{Proceedings of the 2018 Conference of the North {A}merican
  Chapter of the Association for Computational Linguistics: Human Language
  Technologies, Volume 1 (Long Papers)}, pages 756--766, Association for
  Computational Linguistics, New Orleans, Louisiana.

\bibitem[{Yang et~al.(2023)Yang, Xu, Hu, and Dong}]{YANG2023103245}
Yang, Zongbao, Yinxin Xu, Jinlong Hu, and Shoubin Dong. 2023.
\newblock Generating knowledge aware explanation for natural language
  inference.
\newblock \emph{Information Processing \& Management}, 60(2):103245.

\bibitem[{Zhao, Huang, and Ma(2016)}]{zhao-etal-2016-textual}
Zhao, Kai, Liang Huang, and Mingbo Ma. 2016.
\newblock Textual entailment with structured attentions and composition.
\newblock In \emph{Proceedings of {COLING} 2016, the 26th International
  Conference on Computational Linguistics: Technical Papers}, pages 2248--2258,
  The COLING 2016 Organizing Committee, Osaka, Japan.

\end{thebibliography}

\end{document}